\documentclass[journal]{IEEEtran}
\ifCLASSINFOpdf
\else
\fi
\ifCLASSOPTIONcompsoc
 \usepackage[caption=false,font=normalsize,labelfont=sf,textfont=sf]{subfig}
\else
 \usepackage[caption=false,font=footnotesize]{subfig}
\fi
\usepackage[utf8]{inputenc}
\usepackage[mathscr]{euscript}
\usepackage{booktabs}
\usepackage{epsfig}
\usepackage{epstopdf}
\usepackage[usenames, dvipsnames]{color}
\usepackage{bm}
\usepackage{multirow}
\usepackage{graphicx}
\usepackage[linesnumbered, ruled,vlined]{algorithm2e}
\makeatletter
\newcommand*{\rom}[1]{\expandafter\@slowromancap\romannumeral #1@}

\makeatother
\usepackage{amsmath}
\usepackage{amssymb}
\usepackage{amsthm}
\usepackage{mathtools}

\allowdisplaybreaks

\begin{document}
%
\title{Power System Event Identification based on Deep Neural Network with Information Loading \\
\thanks{Disclaimer: this report was prepared as an account of work sponsored by an agency of the United States Government.  Neither the United States Government nor any agency thereof, nor any of their employees, makes any warranty, express or implied, or assumes any legal liability or responsibility for the accuracy, completeness, or usefulness of any information, apparatus, product, or process disclosed, or represents that its use would not infringe privately owned rights.  Reference herein to any specific commercial product, process, or service by trade name, trademark, manufacturer, or otherwise does not necessarily constitute or imply its endorsement, recommendation, or favoring by the United States Government or any agency thereof.  The views and opinions of authors expressed herein do not necessarily state or reflect those of the United States Government or any agency thereof.}
}
%
%
%

\author{Jie Shi,~\IEEEmembership{Member,~IEEE,} Brandon Foggo,~\IEEEmembership{Member,~IEEE,} and Nanpeng Yu,~\IEEEmembership{Senior Member,~IEEE} 
\thanks{J. Shi, B. Foggo, and N. Yu are with the Department
of Electrical and Computer Engineering, University of California, Riverside, CA 92501 USA. e-mail: nyu@ece.ucr.edu}
}

%
%

\markboth{Journal of \LaTeX\ Class Files,~Vol.~14, No.~8, August~2015}%
{Shell \MakeLowercase{\textit{et al.}}: Bare Demo of IEEEtran.cls for IEEE Journals}
%



\maketitle

\begin{abstract}
Online power system event identification and classification is crucial to enhancing the reliability of transmission systems. In this paper, we develop a deep neural network (DNN) based approach to identify and classify power system events by leveraging real-world measurements from hundreds of phasor measurement units (PMUs) and labels from thousands of events. Two innovative designs are embedded into the baseline model built on convolutional neural networks (CNNs) to improve the event classification accuracy. First, we propose a graph signal processing based PMU sorting algorithm to improve the learning efficiency of CNNs. Second, we deploy information loading based regularization to strike the right balance between memorization and generalization for the DNN. Numerical results based on real-world dataset from the Eastern Interconnection of the U.S power transmission grid show that the combination of PMU based sorting and the information loading based regularization techniques help the proposed DNN approach achieve highly accurate event identification and classification results.

\end{abstract}


%
\IEEEpeerreviewmaketitle

\section{Introduction}
\IEEEPARstart{D}{riven} by the need to improve the reliability of the power grid following the 2003 blackout in the Northeastern United States, phasor measurement unit (PMU) usage has experienced exponential growth. Nearly 2,000 PMUs are currently deployed in North America and over 3,000 PMUs are currently commissioned in the Chinese power grid \cite{phadke2018phasor}. By leveraging the fast streaming PMU data, various algorithms have been developed to enhance power system operators' situational awareness.

To further improve system reliability, a highly accurate and automatic event detection and identification algorithm is in critical need. When power system events are correctly detected and classified in a timely manner, appropriate corrective control actions can be taken by system operators or control systems to prevent blackouts. Many power system event detection algorithms using PMU data have been developed. However, very few researchers have explored the event classification problem due to a general lack of access to large amounts of real-world streaming PMU data. By leveraging over 2 years of streaming data from 187 PMUs and over 1,000 events, this paper develops an accurate deep neural network based power system event identification and classification algorithm.

\par The topic of data-driven power system event detection has been studied extensively. We summarize the related literature which can be grouped into five categories. The algorithms in the first category detect power system events by performing spectral analysis such as wavelet transforms \cite{kim2015wavelet, negi2017event}, short-time Fourier transforms \cite{sohn2012event}, graph Fourier transforms \cite{shiGSP}, and self-coherence method \cite{zhou2015} on the PMU data. The algorithms in the second category first develop forecasts of PMU data, then an event is detected if the forecast error exceeds some threshold \cite{zhou2018nonparametric, hannon2019real, xie2014dimensionality}. The third category algorithms monitor the variation of spatial correlations among different PMUs via the correlation coefficient matrix \cite{wu2014real}, the sample covariance matrix \cite{ling2019new}, and the tensor sample covariance matrix \cite{shi2019dimensionality}. A large variation in spatial correlations indicates the occurrence of a power system event. The fourth category of approaches exploits the low-rank property of PMU data during non-event periods. A significant increase in data matrix rank is treated as the sign for a power system event \cite{gao2015missing, hao2018m}. The last group of algorithms use data mining techniques such as matrix profile \cite{shi2019discovering} to detect power system events.
\par Although the sub-field of data-driven event detection has seen tremendous development, the subject of power system event identification and classification has not been fully explored. The insufficient research in this area is mainly due to the lack of access to large-scale labeled real-world PMU data. Most of the existing work leverages just a small amount of real-world synchrophasor data and a limited number of event labels from a restricted class of events. For example, the dataset of reference \cite{li2018real} only contains 32 labeled power system events. Reference \cite{hannon2019real} uses historical data of a single PMU, and reference \cite{dahal2012preliminary} uses just 4 PMUs for its case studies.

\par To overcome the challenges associated with the lack of access to real-world data, some researchers have tried to create and leverage synthetic PMU data \cite{brahma2016real, li2019unsupervised, li2019hybrid}. However, this workaround has its own drawbacks. It is extremely difficult to generate large-scale noisy streaming PMU data during power system events with time-varying spatial temporal correlations similar to that of the real-world data. 

\par Two recent works have managed to collect a relatively large amount of synchrophasor data. Reference \cite{nguyen2015smart} used one-year of historical data from 44 PMUs in the Pacific Northwest. However, the dataset only contains 57 labeled line events, which limits its use for training a general event classifier. Similarly, reference \cite{wang2020frequency} used hundreds of labeled frequency events from the FNET/GridEye system to train a frequency event detection algorithm based on a deep neural network. However, the lack of labels for other event types makes it infeasible to develop a general event identification and classification model.

\par Equipped with two years of data from hundreds of PMUs and over one thousand event labels of different types in the Eastern Interconnection of the U.S. power transmission grid, this paper aims at developing a deep neural network based framework to identify and classify power system events in real time. To deal with high-dimensional PMU input arrays, i.e., tensors, we adopt the convolutional neural networks (CNNs) as the base model. CNNs enable sparse interaction which not only greatly reduces the memory requirements of the model, but also improves its statistical efficiency. To make parameter sharing more effective in the CNN framework, we proposed an innovative graph signal processing (GSP) based PMU sorting algorithm that systematically arranges PMUs in the input tensor. To further improve the event identification and classification accuracy, we propose an information loading based regularization technique to control the amount of information compression between the input layer and the last hidden layer of the deep neural network.

\par The idea of information loading based regularization originates from our previous work \cite{foggo2019improving}. Nevertheless, this work presents several unique innovative algorithm designs and findings. First, the information loading based regularization is still a new theory, which has only been shown to work well with low entropy dataset. Our large-scale experiments in this work show that the technique is still effective when dealing with high entropy data (e.g., PMU data) if the regularization weight is relatively small. Our experiments show that the information loading based regularization performs well with high entropy dataset with a relatively small number of training samples, which is not expected in \cite{foggo2019improving}. The theoretical reason behind this success could be that the deep neural network overly compresses the information between the input layer and the last hidden layer during training session \cite{shwartz2017opening}. Second, in this study we design a neural network structure based on convolutional neural network (CNN), which is completely different from the feed-forward neural network used in \cite{foggo2019improving}. In other words, we extend the application of information loading based regularization from feed-forward neural network to deep CNNs. In addition, we find that the combination of GSP based PMU sorting and information loading based regularization works best in this study. The unique contributions of this work are summarized as follows:

\begin{itemize}
    \item We develop a novel deep neural network structure with information loading based regularization for power system event identification and classification.
    \item We propose a GSP based PMU sorting algorithm to make the parameter sharing scheme more effective in the proposed CNN framework.
    \item Our proposed deep neural network based approach achieves  a high F1 score on the power system event classification task using real-world PMU data.
    \item The representations learned by the deep neural network are interpretable and meaningful.    
\end{itemize}

\par The rest of the paper is organized as follows: Section \ref{framework} presents the problem formulation and overall framework of the proposed deep neural network based power system event identification algorithm. Section \ref{method} provides the technical details of the proposed approach. Section \ref{numericalStudy} validates the proposed power system event identification framework with real-world PMU data. Section \ref{conclusion} states the conclusions.

\section{Problem Formulation and Overall Framework}\label{framework}
In this section, we present the problem formulation and the overall framework of the proposed approach for power system event identification and classification.
The power system event identification problem can be formulated as a statistical classification problem. We develop a deep neural network and train it with historical PMU data with the corresponding power system event labels. When the training process completes, the fitted model servers as an online classifier to identify different power system events using streaming PMU data.
\begin{figure}[!t]
	\centering
	\includegraphics[width=0.48\textwidth]{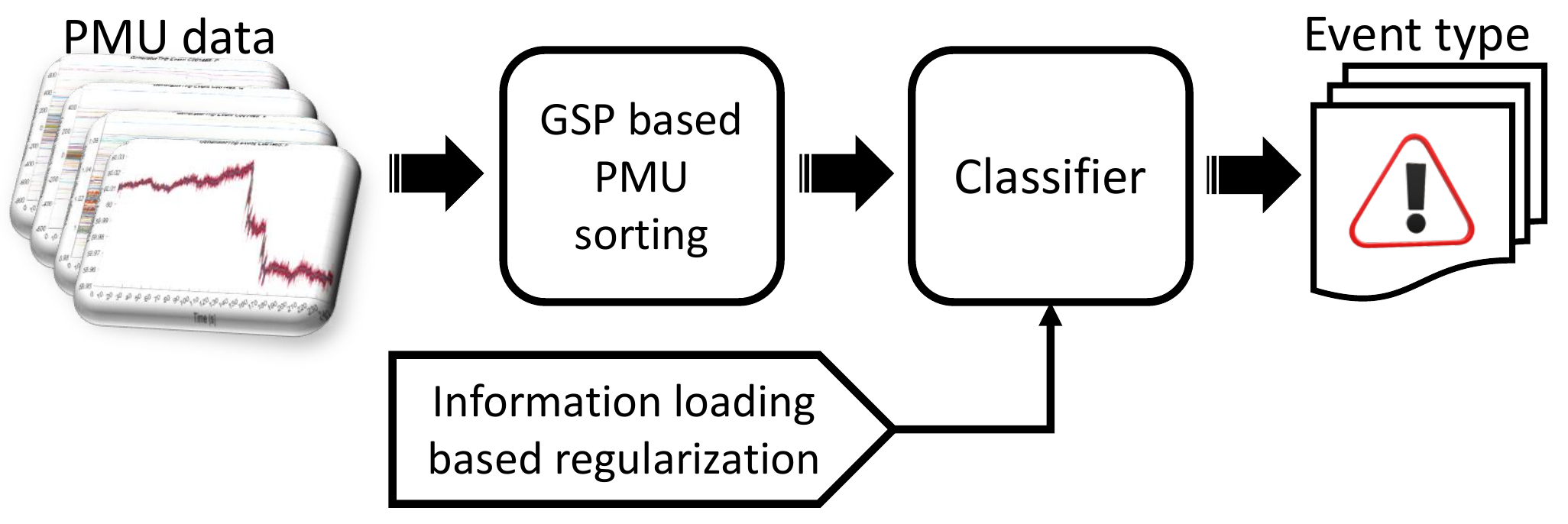}
	\DeclareGraphicsExtensions.
    \caption{Overall framework.}
	\label{overallFramework}
\end{figure}

The overall framework of the proposed approach is shown in Fig. \ref{overallFramework}. It has three key modules: the neural classifier based on CNN, the GSP based PMU sorting algorithm, and the information loading based regularization technique.
\par
\begin{figure}[!t]
	\centering
	\includegraphics[width=0.45\textwidth]{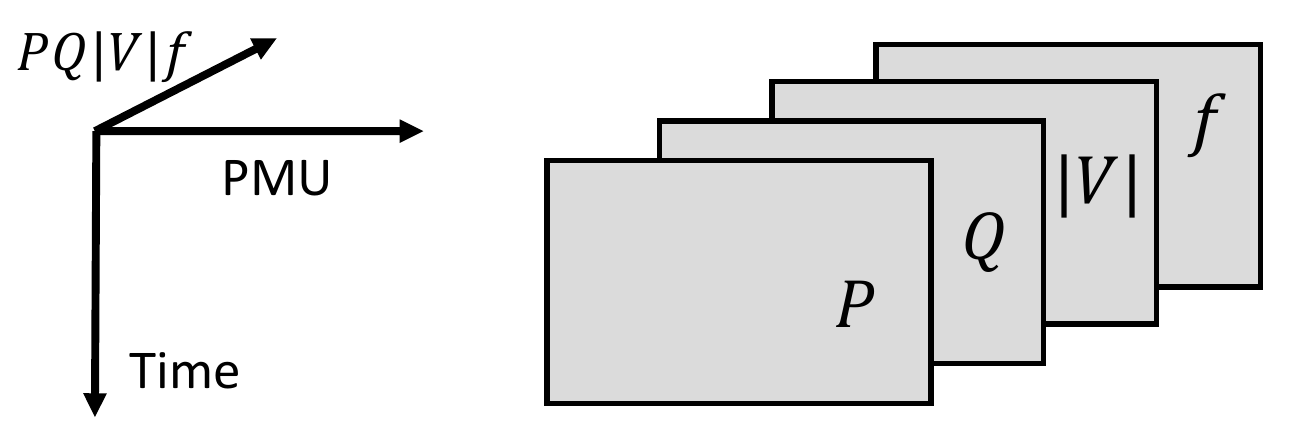}
	\DeclareGraphicsExtensions.
    \caption{Illustration of a $PQ|V|f$ tensor.}
	\label{pqvfTensor}
\end{figure}

 The raw PMU input data array to the proposed algorithm is a 3-dimensional tensor of real power ($P$), reactive power ($Q$), voltage magnitude ($|V|$), and frequency ($f$) measurements from multiple PMUs for a certain time period. The 3 dimensions of the tensor represent time, PMU ID, and $PQ|V|f$ index as illustrated in Fig. \ref{pqvfTensor}. For real-world power grids, the number of deployed PMUs can reach hundreds or thousands.
\par The original PMU data-points are stacked together without any systematic arrangement. Given the reliance of CNNs on exploiting feature localities, this unorganized placement of PMU data is likely to hinder good performance. Arranging PMU time series systematically according to the electrical distance would significantly benefit the CNN based classifier. However, the topology information of the electric grid is not known. Thus, we develop a GSP based PMU sorting module that takes the raw PMU tensor and outputs a sorted PMU tensor which is more compatible with the CNN. In the sorted tensor, PMUs with highly correlated measurements are placed closer to each other. The technical methods used in the GSP based PMU sorting module is presented in Section \ref{methodGSP}.
\par The CNN based classifier module first takes the sorted PMU tensors as inputs and leverage convolution filters in successive layers to transform the inputs into interpretable hidden representations.  The type of power system event of the corresponding time window is then identified by the estimation layer. The overall neural network design is described in Section \ref{Architecture}.

\par One important goal of this work is to achieve higher identification accuracy of power system events by introducing state-of-the-art machine learning techniques. To this end, we employ a novel neural network regularization technique called information loading. The information loading based regularization originates from our earlier work \cite{foggo2019improving}. Its key idea is to boost the classification accuracy of a neural network by adjusting the mutual information between the input features and the learned representations.

\par The information loading based regularization technique is applied in the neural network training process. It estimates the mutual information between the input tensors and the hidden representations, which is added as a penalty term to the typical cross-entropy loss function. The information loading technique is based on the information bottleneck theory \cite{shwartz2017opening} and our recent theoretical results on new bounds for information losses in neural classifiers \cite{foggo2019information}. The theoretical foundation of information losses and technical methods of information loading based regularization will be presented in Section \ref{info}.

\section{Technical Methods}\label{method}
The technical methods of the proposed power system event identification approach are presented in this section. The GSP based PMU sorting algorithm is described in the first subsection, which is followed by an explanation of the information loading based regularization technique. 
We close this section by providing the architecture of the proposed deep neural network for power system event identification.
\subsection{Graph Signal Processing Based PMU Sorting}\label{methodGSP}
In this subsection, we propose a sorting algorithm based on GSP to systematically arrange PMUs in the $PQ|V|f$ tensor. In order to leverage the powerful convolution operation in deep neural network design, we strategically place highly correlated PMUs close to each other. Again, the goal is to determine a useful ordering of PMUs in the input tensor. Let $N$ denote the total number of PMUs and $d_i$ be the index of the $i$th PMU in the ordered input tensor.

\par We can consider this problem as placing PMUs on a straight line. Let $d_i$ be the position of the $i$-th PMU on this line. The positions themselves reflect the sorting of PMUs. Our goal is to find an ordering $\bm{d}=\{d_1,\cdots,d_N\}$ such that the total variance between PMUs is minimized. In this study, we define the total variance as $\frac{1}{2}\sum_{i=1}^N\sum_{j=1}^NW_{ij}\left(d_i-d_j\right)^2$, where $W_{ij}$ represents the absolute value of the correlation coefficient between PMUs $i$ and $j$'s measurements. This total variance represents a weighted sum of squared distances between PMU positions. Due to the weighting scheme, this objective function prefers keeping highly correlated PMUs close to each other.
\par Then the problem of finding $\bm{d}=\{d_1,\cdots,d_N\}$ such that similar PMUs are arranged closer together can be formulated as the following optimization process: 
\begin{align}
\underset{\bm{d}}{\text{minimize}}&\qquad \frac{1}{2}\sum_{i=1}^N\sum_{j=1}^NW_{ij}\left(d_i-d_j\right)^2 \label{obj} \\
\text{subject to}&\qquad \bm{d}^T\bm{d}=1  \label{c1}\\
&\qquad \bm{d}^T\bm{1}=0  \label{c2}
\end{align}
The objective function (\ref{obj}) denotes a correlation-weighted sum of squared distances between PMU positions. Constraints (\ref{c1}) and (\ref{c2}) ensure that the positions of the PMUs in the ordered tensor are centered around the origin and not placed at one single point. This optimization problem is nonlinear and non-convex due to the quadratic equality constraint. 


\par We develop an algorithm to solve this problem efficiently. In the following paragraphs, we solve this optimization problem by exploiting the concepts and methods from graph signal processing.

\par We can envision that all PMUs (vertices) are connected to each other in a complete graph. Let $G=(\mathcal{V}, \mathcal{E})$ be a complete graph, where $\mathcal{V}=\{v_1,\cdots, v_N\}$ is a set of vertices representing the PMUs within a power system. $\mathcal{E}=\{e_{ij}|i\ne j; i,j=1,\cdots,N\}$ is a set of edges where $e_{ij}$ stores the Pearson correlation coefficient between the $i$th and $j$th PMU's measurements. Define $W$ as a weight matrix with diagonal elements being zeros. Its non-diagonal element $W_{ij}$ \textcolor{blue}{is} set as $|e_{ij}|$. Let $L=D-W$ be the graph Laplacian, where the degree matrix $D$ is a diagonal matrix with $D_{ii}=\sum_{j=1}^NW_{ij}$. Then we have the following relationship \cite{stankovic2019vertex}:
\begin{align}
    \bm{d}^TL\bm{d} &=\sum_{i=1}^N\sum_{j=1}^NW_{ij}\left(d_i^2-d_id_j\right) \nonumber \\ &=\frac{1}{2}\sum_{i=1}^N\sum_{j=1}^NW_{ij}\left(d_i^2-d_id_j+d_j^2-d_jd_i\right)^2 \nonumber \\ &=\frac{1}{2}\sum_{i=1}^N\sum_{j=1}^NW_{ij}\left(d_i-d_j\right)^2 \label{gsp2}
\end{align}

By substituting (\ref{gsp2}) and (\ref{c1}) into (\ref{obj}), the objective function (\ref{obj}) can be converted to the Rayleigh quotient $\bm{d}^TL\bm{d}/\bm{d}^T\bm{d}$. The converted unconstrained objective function has a minimum value equal to the smallest eigenvalue of the graph Laplacian $L$, which is $0$. The optimal solution to the unconstrained objective function is the the eigenvector of $L$ that corresponds to the smallest eigenvalue, which is $\bm{1}/\sqrt{N}$. However, (\ref{c2}) rules out this solution by constraining $\bm{d}$ to be orthogonal to $\bm{1}/\sqrt{N}$. Therefore, the solution of \textit{constrained} optimization problem (\ref{obj})-(\ref{c2}) is the eigenvector corresponding to $L$'s second smallest eigenvalue. Thus, the PMUs measurements should be sorted according to the optimal solution $\bm{d}$ in an ascending order. Algorithm \ref{algGSP} summarizes the procedures of the proposed GSP based PMU sorting approach.
\begin{algorithm}[!t]
	Obtain the Pearson correlation coefficients between PMUs\;
	Construct weight matrix $W$ and Laplacian graph $L$\;
	Take eigendecomposition of $L$\;
	Sort PMUs according to the eigenvector corresponding to the second smallest eigenvalue of $L$\;
	\caption{GSP based PMU sorting algorithm}
	\label{algGSP}	
\end{algorithm}

\subsection{Information Loading based Regularization}\label{info}
The motivation and algorithm for the information load based regularization technique are presented in this subsection. Information loading is a regularization technique first proposed in our previous work \cite{foggo2019improving}. It is motivated by the information bottleneck theory \cite{shwartz2017opening} and recent theoretical results about information losses of neural classifiers \cite{foggo2019information}. The information loading based regularization technique controls the amount of information compression between the input layer and the last hidden layer of a deep neural network. First, we briefly review the theory of information losses and then present the information loading algorithm.
\subsubsection{Information Losses}
\begin{figure}[!t]
	\centering
	\includegraphics[width=0.48\textwidth]{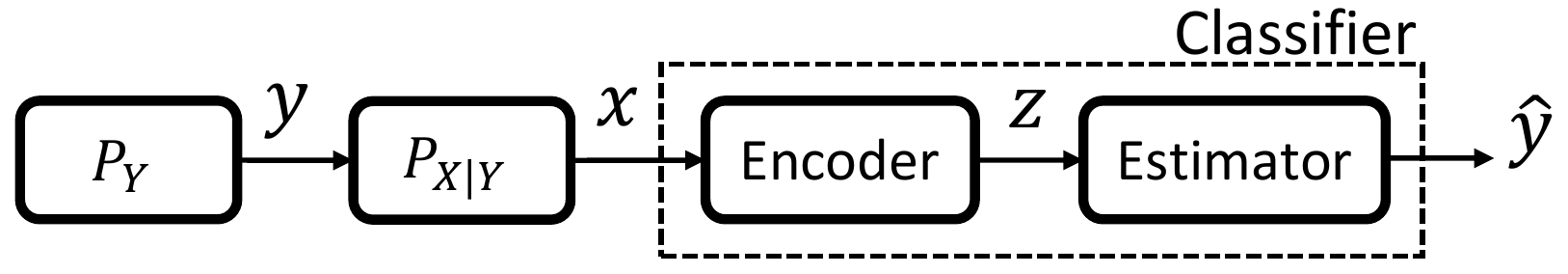}
	\DeclareGraphicsExtensions.
    \caption{Flowchart of a general classification model.}
	\label{flowchart}
\end{figure}
A general classification model can be represented as a Markov chain as shown in Fig. \ref{flowchart}. $Y$ and $\hat{Y}$ denote the real class label and the estimated label. $X$ denotes the input features. $Z$ denotes the learned representation (in our case, the last hidden layer of a deep neural network). $y$, $\hat{y}$, $x$, and $z$ are instances of the random variables $Y$, $\hat{Y}$, $X$, and $Z$, respectively. The classifier consists of two parts: an encoder that models $P_{Z|X}$ and an estimator that models $P_{\hat{Y}|Z}$. 

To achieve better classification accuracy, we want to learn a representation $Z$ which has a high mutual information $I(Y;Z)$ with the class label. But in practice, the exact value of $I(Y;Z)$ can not be obtained from the training data. Instead, we must resort to using an estimate, denoted $\hat{I}(Y;Z)$, which has the following form:
\begin{align}
    \hat{I}(Y;Z)=\mathbb{E}\left[\log_2\frac{\hat{P}_{ZY}}{\hat{P}_{Z}\hat{P}_{Y}}\right]
\end{align}
where $\hat{P}_{ZY}$, $\hat{P}_{Z}$, and $\hat{P}_{Y}$ denote the estimates of $P_{ZY}$, $P_{Z}$, and $P_{Y}$ based on the training samples.

Let $Z^*$ and $\tilde{Z}$ be the representations that, respectively, maximize $I(Y;Z)$ and $\hat{I}(Y;Z)$ with the constraint $I(X;Z)=C$, where $C$ is a constant that defines the complexity of encoder. In other words, $Z^*$ is produced by the optimal encoder when we have perfect knowledge of $P_{XY}$. $\tilde{Z}$ is produced by the optimal encoder when we have partial knowledge of $P_{XY}$.

Now, we can define the term ``information losses'', which is strongly related to `minimal classification error' \cite{foggo2019improving}:
\begin{align}
    I_{loss} = |I(Y;Z^*)-I(Y;\tilde{Z})| \label{infoloss}
\end{align}
The upper bound of the ``information losses'' is given by \cite{foggo2019information}:
\begin{align}
    I_{loss} \le 2\left(\delta_{\hat{P}}I(X;\tilde{Z})+h_2(\delta_{\hat{P}})\right)+\epsilon \label{infoBoundEq}
\end{align}
where $h_2(\cdot)$ denotes the binary entropy function. $\delta_{\hat{P}}$ is the conditional total variation of $\hat{P}_{Y|X}$ from $P_{Y|X}$:
\begin{align}
    \delta_{\hat{P}}=\frac{1}{2}\mathbb{E}_{P_X}\left[\sum_{y\in \mathscr{Y}}\left|P_{Y|X}(y|x)-\hat{P}_{Y|X}(y|x)\right|\right]
\end{align}
$\mathscr{Y}$ is the set of all class labels. Note that $\delta_{\hat{P}}$ has an upper bound that only depends on the given training data \cite{foggo2020maximum}. 
\par The upper bound of information losses (\ref{infoBoundEq}) has been shown to be reasonably tight across many testing datasets \cite{foggo2019information}. It will serve as the cornerstone in the information loading based regularization technique.

\begin{figure}[!t]
    \centering
    \subfloat[Low entropy data.]{\includegraphics[width=0.23\textwidth]{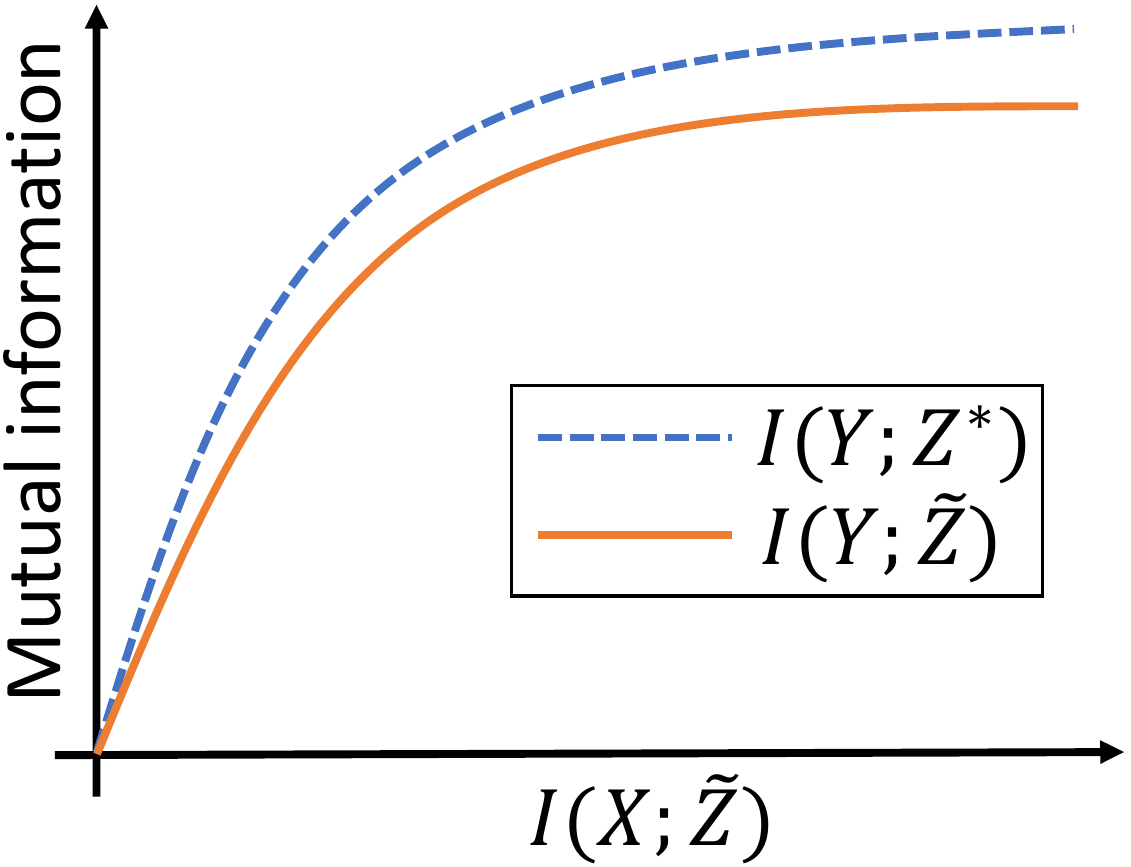}%
    \label{lowEntropy}}
    \hfil
    \centering
    \subfloat[High entropy data.]{\includegraphics[width=0.23\textwidth]{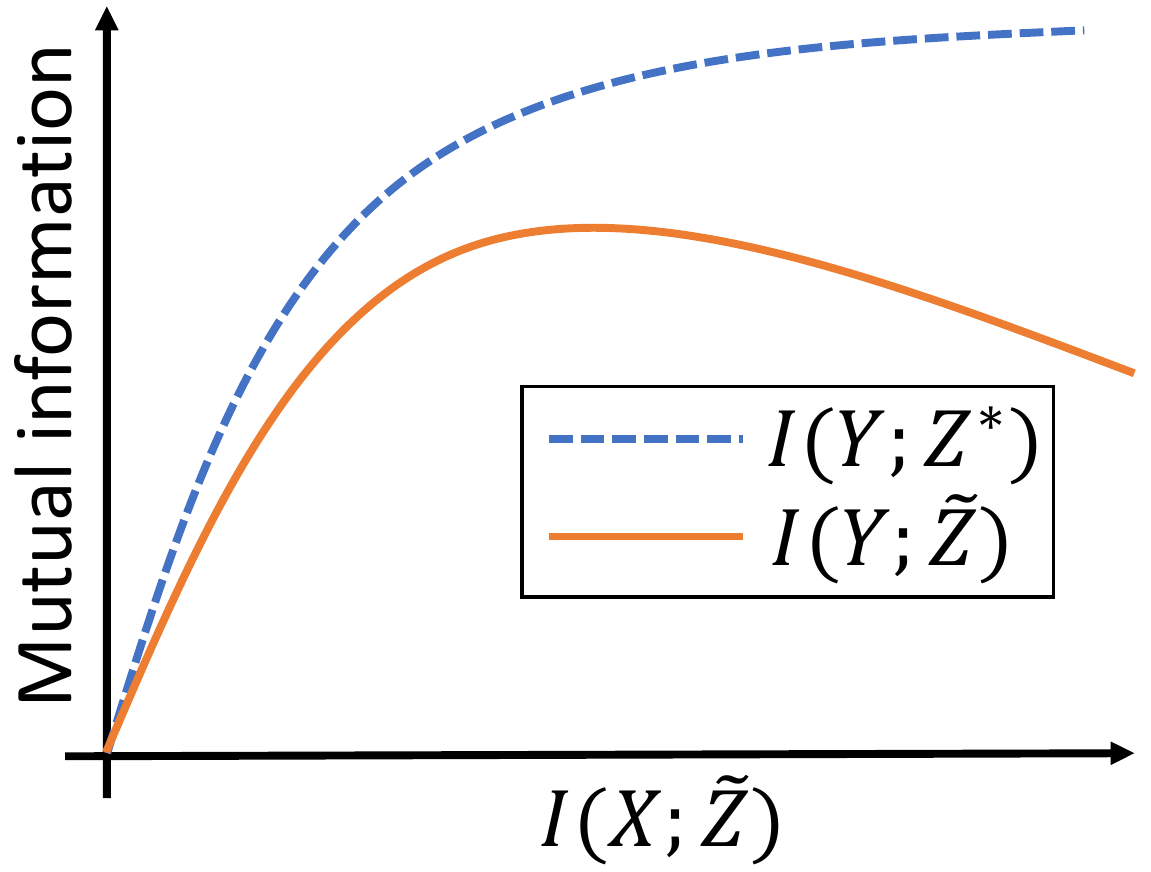}%
    \label{highEntropy}}
    \caption{Mutual information bounds for low and high entropy feature space.}
    \label{infoBoundPlt}
\end{figure}
\subsubsection{Information Loading}
Based on the theoretical results of information losses, we can visualize the mutual information bounds for typical datasets. Suppose $I(Y;Z^*)$ is given, we can draw the lower bound of $I(Y;\tilde{Z})$ based on (\ref{infoloss}) and (\ref{infoBoundEq}). It has been shown that this lower bound behaves differently depending on the entropy level of the input feature space \cite{foggo2019information} as illustrated in Fig. \ref{infoBoundPlt}.

For low entropy input features, the lower bound keeps increasing as the complexity of the encoder $I(X;\tilde{Z})$ increases. In this case, we can safely boost the mutual information between the input and the hidden representation $I(X;\tilde{Z})$ to achieve higher $I(Y;\tilde{Z})$ and classification accuracy. However, we need to be careful about adjusting $I(X;\tilde{Z})$ for high entropy input data. If the representation becomes too complex, the performance of the classifier will deteriorate. 

Previous research \cite{shwartz2017opening} shows that during the training process, deep neural networks often compress information between the input layer and the last hidden layer, i.e., reducing $I(X;\tilde{Z})$. Thus, we can penalize the compression of information by augmenting the typically cross-entropy loss function of classification model with $-\beta \hat{I}(X;Z)$, where $\hat{I}(X;Z)$ is an estimate of $I(X;Z)$. This will help $I(Y;\tilde{Z})$ approach its peak.

The selection of hyperparameter $\beta$ depends on the entropy of the input feature space. As explained before, for a low entropy feature space, $\beta$ could be selected to be a larger value (in accordance with the left hand side of Fig. \ref{infoBoundPlt}). However, for a high entropy feature space, a very large $\beta$ may lead to performance degradation (in accordance with the right hand side of Fig. \ref{infoBoundPlt}). 

\subsection{Neural Network Design for System Event Identification} \label{Architecture}
\begin{figure}[!t]
	\centering
	\includegraphics[width=0.48\textwidth]{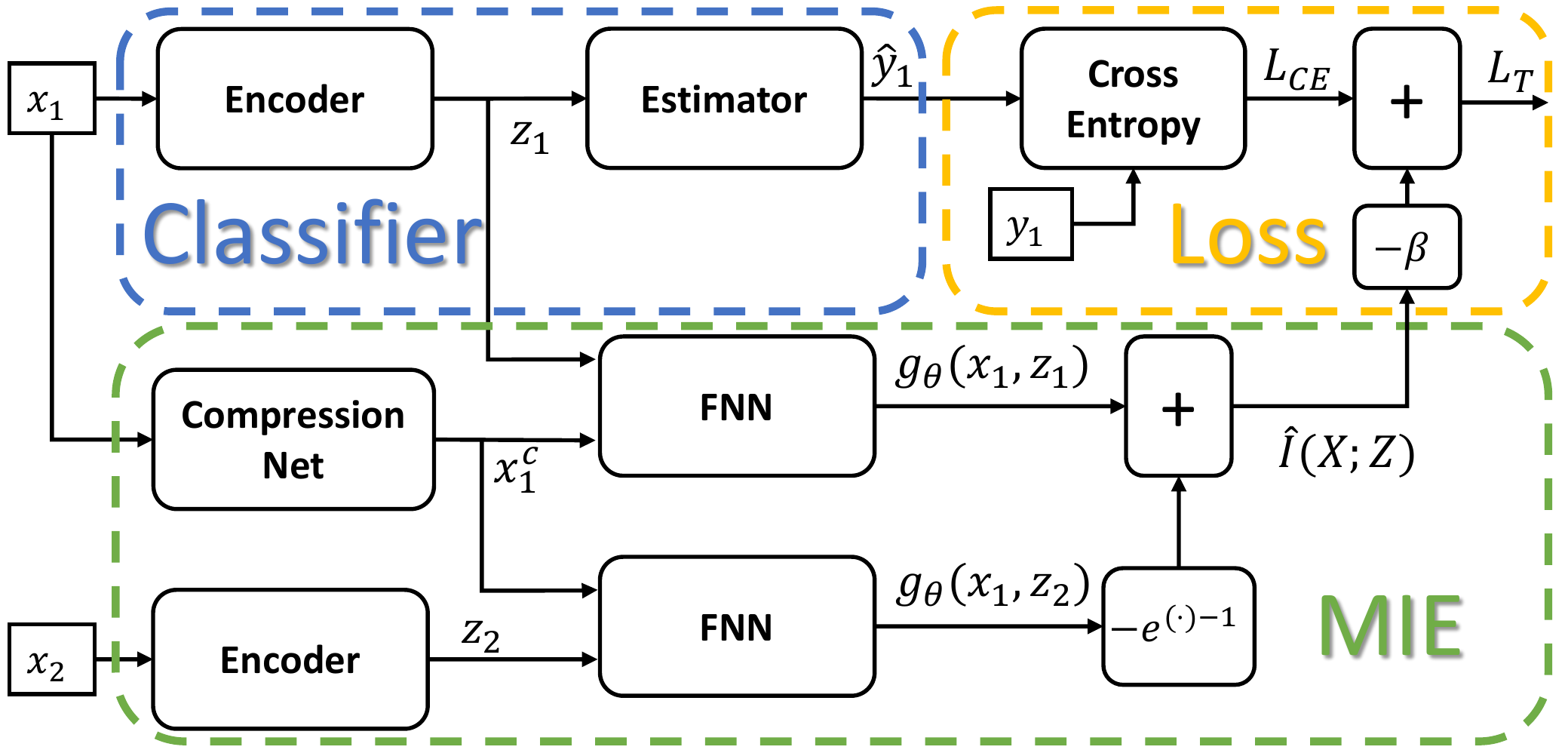}
	\DeclareGraphicsExtensions.
    \caption{The overall neural network architecture.}
	\label{netStructure}
\end{figure}
In this subsection, we present the neural network design for power system event identification using PMU data. The overall neural network architecture of the proposed solution is shown in Fig. \ref{netStructure}. It is composed of three main modules: the classifier model, the mutual information estimator (MIE), and the information loaded loss function. The design for each of the three modules is described in detail below.
\subsubsection{The classifier model}
The classifier is an essential module of the overall design. Both the MIE and the information loaded loss function modules are built to facilitate the training of the classifier. Once trained, the classifier will serve as a standalone unit to identify power system events based on streaming PMU data.
\par The proposed classifier module contains two components: the encoder and the estimator. The encoder transforms the input features into representations. The representations are then fed into the estimator, which produces the final power system event identification results. The input $PQ|V|f$ tensors are arranged based on the GSP based PMU sorting algorithm described in Section \ref{methodGSP}. Since the input $PQ|V|f$ tensors have structures similar to the images, we decided to adopt a widely-used deep convolutional neural network (CNN), ResNet-50 \cite{He_2016_CVPR} as a key building block of the encoder. ResNet and its extensions have achieved great success across various applications, making them one of the most popular deep CNN families in the machine learning community. In this work, the encoder is built by connecting ResNet-50 (excluding the output layer) to a dense layer of 10 neurons. The estimator is designed as one dense layer with the softmax activation function.
\subsubsection{Mutual Information Estimator}
The mutual information estimator is built to provide an estimate of mutual information between the input feature $X$ and the representation $Z$. As discussed in Section \ref{info}, we need to tune $I(X;Z)$ to improve the classification accuracy. To achieve this goal, the estimate $\hat{I}(X;Z)$ will be fed into the information loaded loss function module as a regularization term.
\par The proposed mutual information estimator is inspired by MINE-$f$ introduced in \cite{belghazi2018mutual}. By definition, $I(X;Z)$ can be written in the KL-divergence form:
\begin{align}
    I(X;Z) = D_{KL}(P_{XZ}||P_X\otimes P_Z) \label{klInfo}
\end{align}
where $P_X\otimes P_Z$ denotes the product of the marginal distributions $P_X$ and $P_Z$.  The $f$-divergence representation developed by \cite{nguyen2010estimating, nowozin2016f} provides a lower bound of (\ref{klInfo}):
\begin{align}
    D_{KL}(P_{XZ}||P_X\otimes P_Z)=  \sup_{g\in \mathscr{G}}\mathbb{E}_{P_{XZ}}\left[g\right]-\mathbb{E}_{P_X\otimes P_Z}\left[e^{g-1}\right] \label{klXZ}
\end{align}
where $\mathscr{G}$ denotes the set of all possible mappings. In this study, we parameterize $g$ using a deep neural network $g_{\bm{\theta}}$. As shown in Fig. \ref{netStructure}, $g_{\bm{\theta}}$ is composed of two parts: a compression net and a feed-forward neural network (FNN). The compression net is used to compress the input $PQ|V|f$ tensor into a low-dimensional representation. Specifically, we build the compression net by connecting a lightweight deep CNN called MobileNetV2 \cite{Sandler_2018_CVPR} (excluding the output layer) to a dense layer of 10 neurons. The FNN has two layers of neurons with dimensions of 200 and 1. 
\par Based on (\ref{klInfo}) and (\ref{klXZ}), we can build an estimate of $I(X;Z)$ by drawing samples from the input feature space and the representation as follows:
\begin{align}
    \hat{I}(X;Z) = \frac{1}{B_I}\sum_{i=1}^{B_I}g_{\bm{\theta}}(\bm{x}_{1,i},\bm{z}_{1,i})+\frac{1}{B_I}\sum_{i=1}^{B_I}e^{g_{\bm{\theta}}(\bm{x}_{1,i},\bm{z}_{2,i})-1}
\end{align}
where $B_I$ denotes the training batch size. $\bm{x}_{1,i}$ and $\bm{z}_{1,i}$ are samples drawn from the joint distribution $P_{XZ}$. $\bm{z}_{2,i}$ is a sample drawn from the marginal distribution $P_Z$. During the training session, $\bm{z}_{1}=[\bm{z}_{1,1},\cdots,\bm{z}_{1,B_I}]$ and $\bm{z}_{2}=[\bm{z}_{2,1},\cdots,\bm{z}_{2,B_I}]$ are obtained by feeding two independent input batches $\bm{x}_{1}=[\bm{x}_{1,1},\cdots,\bm{x}_{1,B_I}]$ and $\bm{x}_{2}=[\bm{x}_{2,1},\cdots,\bm{x}_{2,B_I}]$ into the encoder (See Fig. \ref{netStructure}).

\subsubsection{Information loaded loss function}
The loss function $L_T$ of the proposed power system event identification model contains two components:
\begin{align}
    L_T= L_{CE}-\beta \hat{I}(X;Z)
\end{align}
The first component $L_{CE}$ is the typical cross-entropy loss function, which is often used for training classifiers. $L_{CE} = -\sum_{i=1}^{B_I}y_i\cdot\log\hat{y}_i$, where $y_i$ and $\hat{y}_i$ denote the true label and the estimated label of input sample $\bm{x}_{1,i}$, respectively.

The second component of the information loaded loss function, $-\beta \hat{I}(X;Z)$, penalizes the compression of information between the input feature and the representation of the deep neural network. The hyperparameter $\beta$ regularizes the mutual information $I(X;Z)$ to boost the performance of the neural classifier. The Adam optimizer \cite{kingma2014adam} is adopted to minimize the information loaded loss function during training. 
\section{Numerical Study} \label{numericalStudy}
In this section, we validate our proposed power system event identification algorithm using a large-scale real-world PMU dataset. First, we briefly discuss the data source and the data preprocessing steps. Then, we present the GSP based PMU sorting results for our PMU dataset. Next, we explain how to address the class imbalance issue with data augmentation. Finally, we carry out an ablation study to quantify the benefits of the GSP based PMU sorting technique and the information loading based regulation method in improving the power system event identification performance.
\subsection{Data Source}
\begin{figure}[!t]
	\centering
	\includegraphics[width=0.5\textwidth]{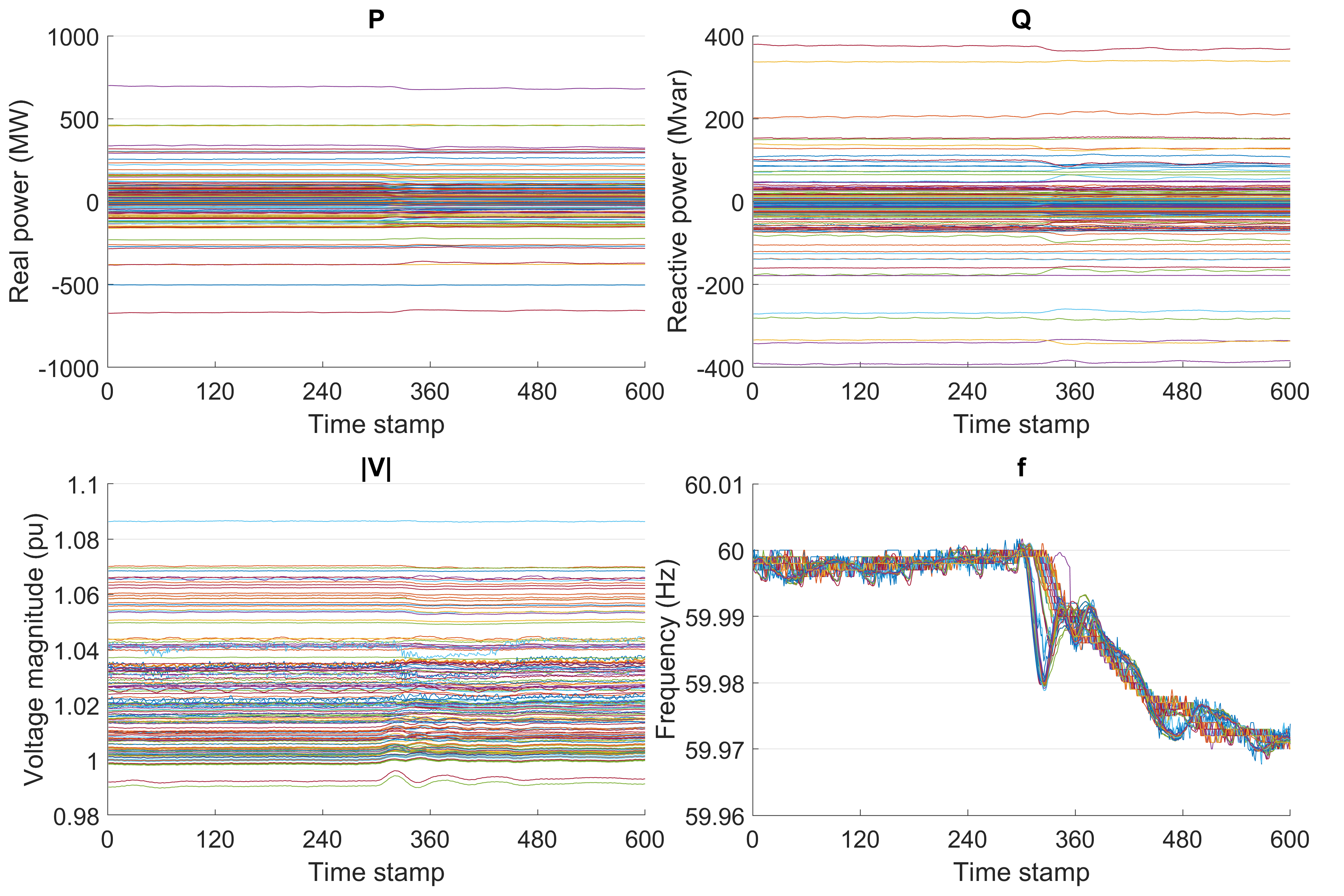}
	\DeclareGraphicsExtensions.
    \caption{$PQ|V|f$ tensor of a sample generator tripping event.}
	\label{sampleGenEvent}
\end{figure}
The dataset comprises two years of PMU data from the Eastern Interconnection of the continental U.S. transmission grid. The PMU measurements are initially collected by electric utility companies and regional system operators (RTOs). The dataset is then compiled by the Pacific Northwest National Laboratory. Both PMU locations and event locations are proprietary information that is unavailable throughout this study.
The raw data from 187 PMUs include measurements of frequency and positive sequence voltage/current magnitudes and phase angles.

\par We convert the raw readings from the PMUs into the corresponding positive sequence real power ($P$), reactive power ($Q$), voltage magnitude ($|V|$), and frequency ($f$) data arrays, i.e., the $PQ|V|f$ tensors. The raw data include 1,147 labeled power system events and 120 non-events. These labels are created by the domain experts from electric utilities and RTOs, which are not publicly available. The time span of each labeled event data sample is 20 seconds, with the event starting time in the middle of the window except for the oscillation events, which cover the entire 20-second windows. 8 PMUs are removed from the analysis due to prevalent bad data. The reporting frequency of the PMU data is 30 Hz. Thus, the dimensionality of the input $PQ|V|f$ tensors is $[600, 179, 4]$, where $600=20\times30$ corresponds to the number of time stamps in the 20-second window and 179 is the number of valid PMUs.
\par Four types of labels are provided for the $PQ|V|f$ tensors. There are 120 \textit{Non-events}, 825 \textit{Line-events}, 84 \textit{Generator-events}, and 118 \textit{Oscillation-events}. The $PQ|V|f$ tensor of a sample generator tripping event is depicted in Fig. \ref{sampleGenEvent}. The \textit{Non-event} class corresponds to time periods without any observable power system events. The \textit{Line-event} class includes line tripping events. The \textit{Generator-event} class includes generator tripping events. The \textit{Oscillation event} class consists of power oscillation events. 
The total number of events is high because the data is collected from the entire Eastern Interconnection in the United States, covering a vast geographical area from the Midwest to the East Coast. The Eastern Interconnection contains more than 6,800 generators, 42,000 buses \cite{overbye2004power}, 31 balancing authorities, and 36 states \cite{hoff2016us} of the U.S. Note that our dataset is severely imbalanced due to the relatively large number of line events. This issue will be addressed by the data augmentation process in Subsection \ref{dataAugSec}. 

\subsection{Bad Data Detection and Missing Value Replacement}
\begin{figure}[!t]
	\centering
	\includegraphics[width=0.45\textwidth]{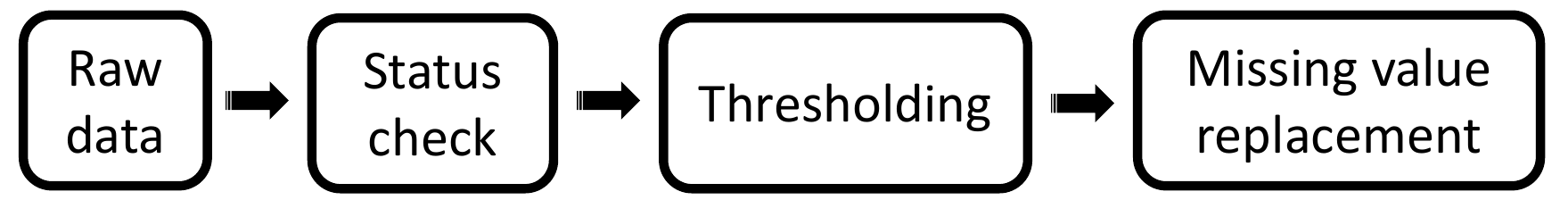}
	\DeclareGraphicsExtensions.
    \caption{Three procedures for missing and bad reading replacement.}
	\label{missingValueReplacement}
\end{figure}
Missing and bad readings are common in real-world PMU datasets. To address this issue, we design three procedures to detect and replace missing and bad readings. These three procedures are status check, thresholding, and missing value replacement (Fig. \ref{missingValueReplacement}).
\\ \textbf{Status check:} The raw PMU readings are accompanied by status flags indicating the conditions of PMUs. According to the IEEE standard for synchrophasor data transfer for power systems (IEEE C37.118.2-2011), the last two bits of status flag indicate the following scenarios:
\begin{itemize}
\item `00': Good measurement data, no error;
\item `01': No data available;
\item `10': PMU in test mode or absent data tag have been inserted;
\item `11': PMU error;
\end{itemize}
These status flags can help us filter out the bad readings when PMUs were malfunctioning or in test mode.
\\ \textbf{Thresholding:} We can still encounter unrealistic readings after the status check. To further filter out bad readings, we introduce a list of ranges on each variable based on domain knowledge. The reading is considered as bad if it falls into one of the following unrealistic ranges:
\begin{itemize}
	\item Voltage magnitude: $(-\infty,0)\cup(1.5 \text{p.u.},+\infty)$
	\item Voltage angle:  $(-\infty,-180^{\circ})\cup(+180^{\circ},+\infty)$
	\item Current magnitude:  $(-\infty,0)\cup(10\text{kA},+\infty)$
	\item Current angle:  $(-\infty,-180^{\circ})\cup(+180^{\circ},+\infty)$
	\item Frequency:  $(-\infty,59\text{Hz})\cup(61\text{Hz},+\infty)$
\end{itemize}
\textbf{Missing value replacement:} In this study, we treat all the 
aforementioned bad readings as missing. The proposed 
missing value replacement algorithm consists of two stages. In the first stage, we mark a PMU as `\emph{NA}' if it has consecutive missing readings longer than one second in the corresponding $PQ|V|f$ tensor. All the readings are assumed to be missing for \emph{NA} PMUs. For the rest of PMUs, we replace their missing values using a subspace estimation approach described in \cite{gao2015missing}. In the second stage, we fill the missing values of \emph{NA} PMUs with the readings of non-\emph{NA} PMUs that have the highest level of correlations. Note that the correlation level between any two PMUs can be approximated by calculating their Pearson correlation coefficient from $PQ|V|f$ tensors during periods without missing values.

\subsection{GSP Based Sorting Result}
\begin{figure}[!t]
	\centering
	\includegraphics[width=0.45\textwidth]{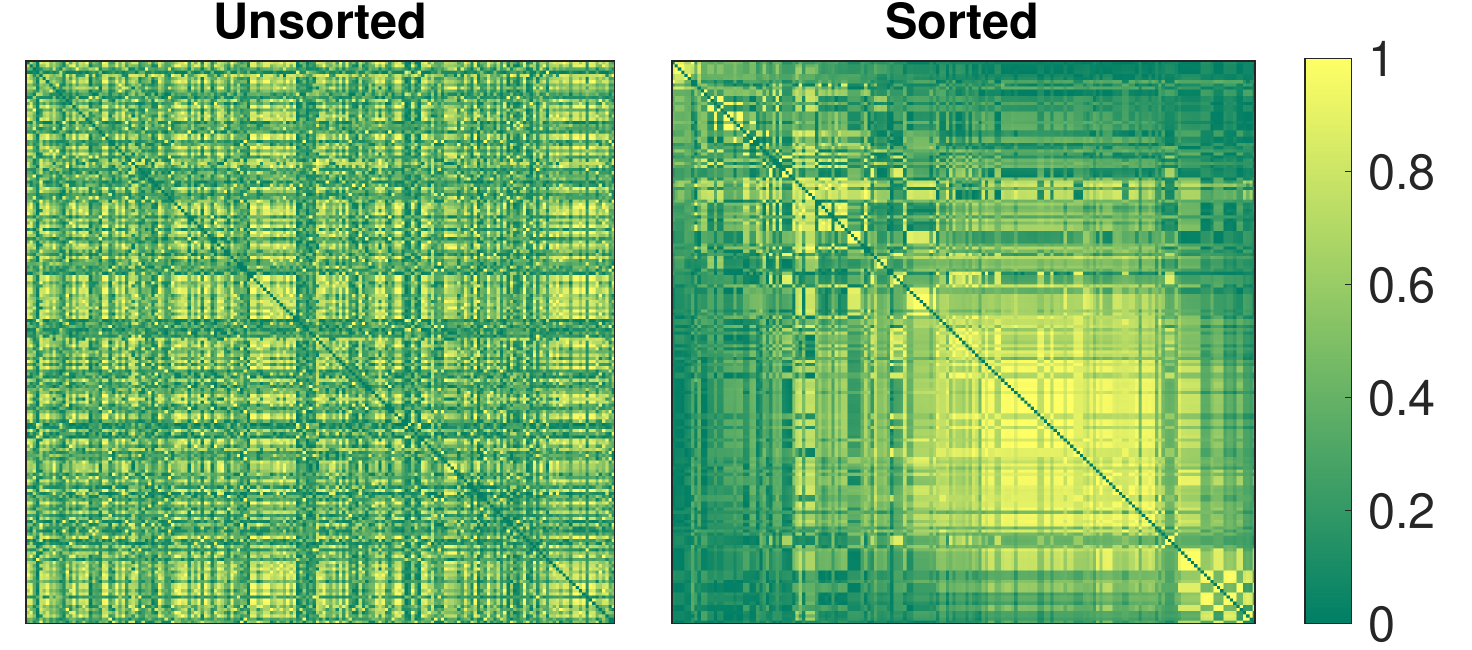}
	\DeclareGraphicsExtensions.
    \caption{Weight matrices of the original and the sorted PMU sequence.}
	\label{unsortSortCorrMtx}
\end{figure}

We calculate the Pearson correlation coefficient for each pair of PMUs. The entries of weight matrix $W$, which quantify the correlation between PMUs, are derived by taking the absolute values of the corresponding correlation coefficients. Fig. \ref{unsortSortCorrMtx} compares the weight matrices of the original PMU sequence and the sorted PMU sequence. As shown in the figure, highly correlated PMUs are placed much closer to each other after the GSP based PMU sorting. It will be shown in Section \ref{performance} that the GSP based PMU sorting technique facilitates the kernel learning in CNN and significantly improves the event classification performance.

\subsection{Data Augmentation}\label{dataAugSec}
\begin{figure}[!t]
	\centering
	\includegraphics[width=0.46\textwidth]{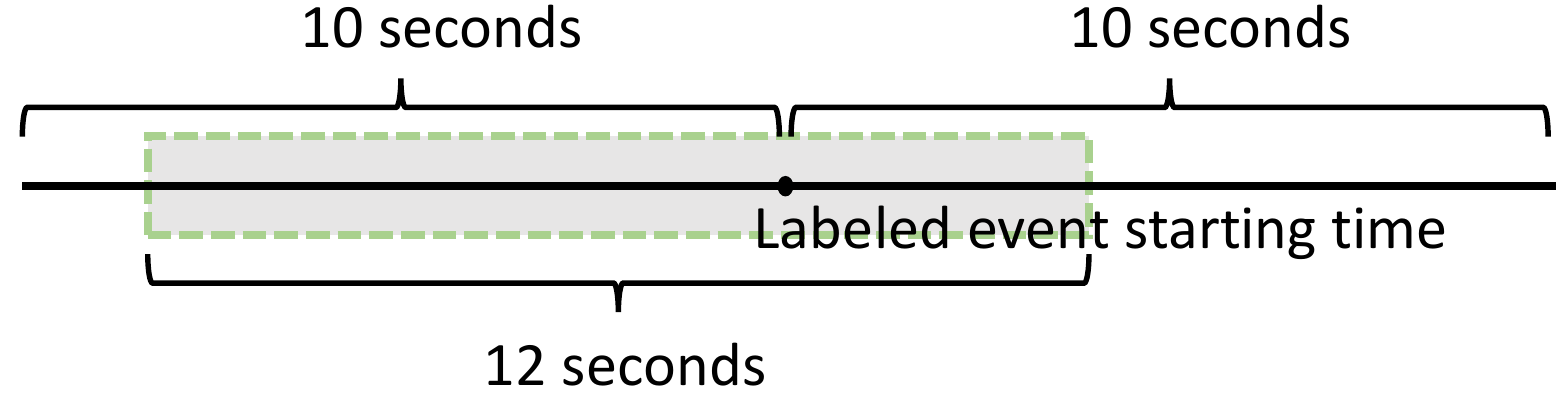}
	\DeclareGraphicsExtensions.
    \caption{Illustration of sub-tensor sampling.}
	\label{dataAug}
\end{figure}

\begin{table}[!t]
\renewcommand{\arraystretch}{1}
\caption{Distribution of $PQ|V|f$ Snapshots} 
\label{dataDistSnap}%
	\setlength{\tabcolsep}{2pt}	
	    \centering
		\begin{tabular}{ccccc}
			\toprule
			 \textbf{Class} & \textit{Non-event}  & \textit{Line-event} & \textit{Generator-event} & \textit{Oscillation-event} \\
			\midrule		
			\textbf{\# of snapshots} &  720  & 825  & 756  & 708 \\
			\bottomrule
		\end{tabular}%
\end{table} 
Two issues exist in our dataset. First, the original $PQ|V|f$ tensors are severely imbalanced due to the relatively large number of line events. The class imbalance can result in over-classification of the majority group due to biased prior distribution \cite{johnson2019survey}. Second, the event starting time stamps for \textit{Line-event} and \textit{Generator-event} samples are always located in the middle of the time windows. In other words, the event signatures consistently appear near the center of the corresponding $PQ|V|f$ tensors, leading to a biased distribution of event timing. Data augmentation is thereby introduced to address these two issues.
\par The core idea of the proposed data augmentation technique is straightforward: we sample sub-tensors from the original $PQ|V|f$ tensors independently and uniformly. Fig. \ref{dataAug} illustrates that a 12-second sub-tensor is sampled uniformly from a given $PQ|V|f$ tensor. Hereafter we will refer to the sampled sub-tensors $PQ|V|f$ as snapshots. It is worth noting that the time range of $PQ|V|f$ snapshots should be large enough to capture the low-frequency power oscillations according to Nyquist–Shannon sampling theorem. In this study, we set this time range to be 12 seconds. 
\par To address the two issues associated with imbalanced dataset, we sample 6, 1, 9, and 6 $PQ|V|f$ snapshots independently and uniformly from each $PQ|V|f$ tensor in \textit{Non-event}, \textit{Line-event}, \textit{Generator-event}, and \textit{Oscillation-event} category, respectively. This procedure creates a relatively balanced dataset as shown in Table \ref{dataDistSnap}. Meanwhile, the event starting time stamps are no longer fixed in the middle of the input tensors, resulting in increased diversity of the event timing.

\subsection{Classification Performance} \label{performance}
\begin{figure}[!t]
    \centering
    \subfloat[Performance comparison on cross-validation.]{\includegraphics[width=0.45\textwidth]{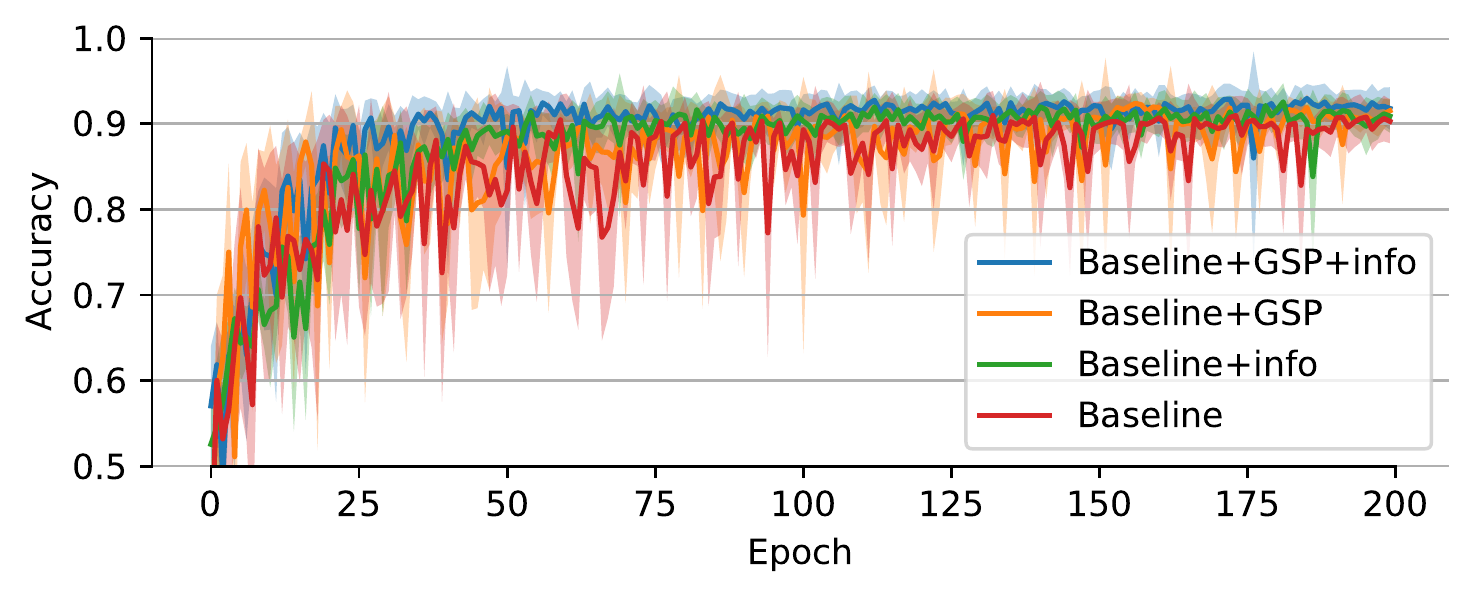}%
    \label{accVal}}
    \hfil
    \centering
    \subfloat[Performance comparison on testing dataset.]{\includegraphics[width=0.45\textwidth]{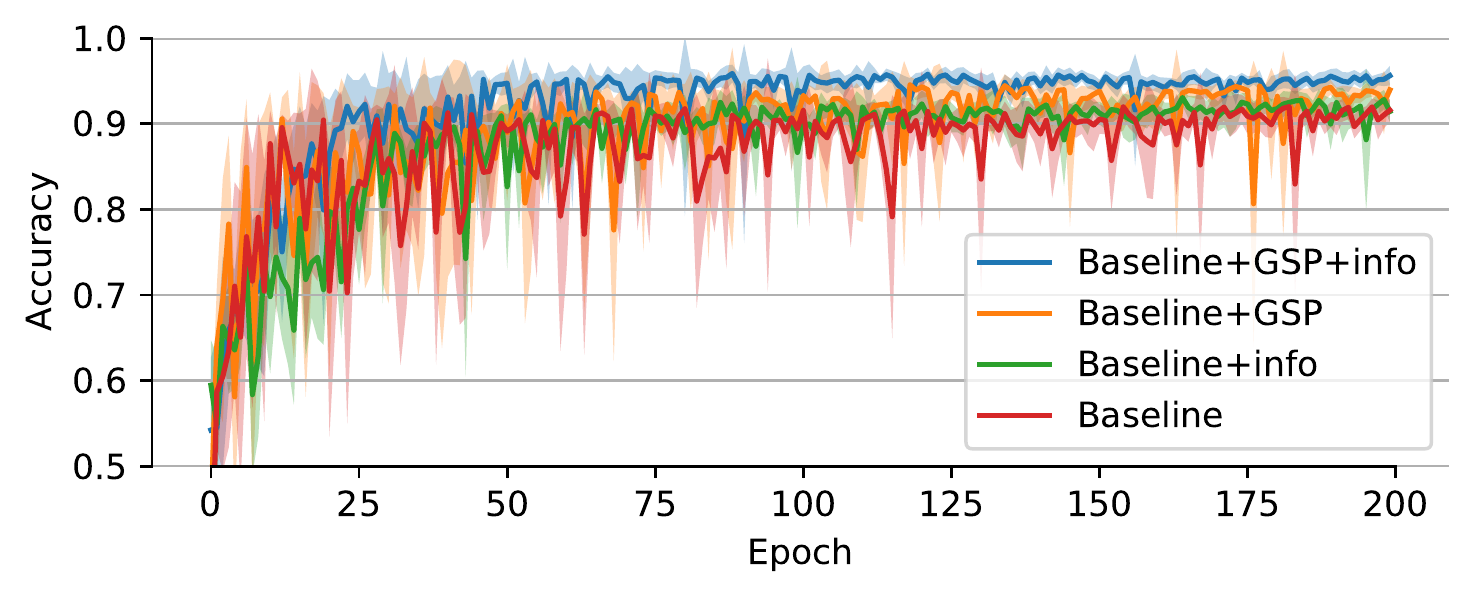}%
    \label{accTest}}
    \caption{Power system event classification accuracy comparison among four different models.}
    \label{classficationResults}
\end{figure}
\renewcommand{\arraystretch}{1}
\begin{table}[!t]
	\setlength{\tabcolsep}{10.5pt}	
	\caption{Average F1 Scores with Different Values of $\beta$} 
	    \centering
		\begin{tabular}{ccccccc}
			\toprule
			  $\beta$ & 0.01 & 0.05 & 0.1 & 0.6 & 1 \\
			\midrule		
			F1 score  & 0.903 & 0.925  & 0.926  & 0.907   & 0.705 \\
			\bottomrule
		\end{tabular}%
	\label{f1scoreCrossVal}%
\end{table} 
We evaluate the performance of the proposed deep neural network based power system event classification algorithm by quantifying its classification accuracy and F1 score on the real-world PMU dataset. We perform an ablation study to tease apart which component(s) of the proposed algorithm are most important for its success. To achieve this goal, we evaluate four methods. The first method directly employs a powerful CNN architecture, ResNet50, without the GSP based PMU sorting or the information loading based regularization techniques. We call the first method \textit{baseline}. The second method represents the baseline CNN combined with information loading based regularization, which is named \textit{baseline+info}. The third method represents the baseline CNN combined with GSP based PMU sorting, which is named \textit{baseline+GSP}. The fourth method is the proposed approach that includes both GSP based PMU sorting and information loading based regularization. We name it \textit{baseline+GSP+info}.

The input $PQ|V|f$ snapshots are divided into two sets: the training dataset (80\%) and the testing dataset (20\%). It is worth noting that the training $PQ|V|f$ snapshots and the testing $PQ|V|f$ snapshots are sampled from different $PQ|V|f$ tensors. Thus, there is no data leakage between the training dataset and the testing dataset. 
\par To achieve a reasonable amount of information compression through the proposed deep neural network, we perform cross-validation on the training dataset to identify an appropriate $\beta$. Specifically, we split the training dataset into 10 subsets and train the neural network for 10 rounds. In each round, 9 subsets are used for training and the other subset is used for validation. This is called 10-fold cross-validation. The settings of each training session are provided as follows. The total number of training epochs is 200. The size of the training batch is 16. The learning rate of the Adam optimizer is selected to be 0.001. We adopt z-score scaling on $P$, $Q$, $|V|$, and $f$ of each PMU in the input $PQ|V|f$ snapshots. 

\par The average F1 scores of the power system event classification results of the \textit{baseline+GSP+info} method under the cross-validation setup is reported in Table \ref{f1scoreCrossVal}. As shown in the table, $\beta=0.1$ achieves the best result in the cross-validation setup. Thus, the hyperparameter $\beta$ is selected to be 0.1. The average accuracy over 10 rounds of four different methods as the training session proceeds are reported in Fig. \ref{accVal}. The proposed method \textit{baseline+GSP+info} performs the best in terms of validation accuracy. The results show that the combination of GSP based PMU sorting and the information loading based regularization is capable of boosting the performance of the baseline model. It is worth noting that there is no significant accuracy difference between \textit{baseline+info} and \textit{baseline}. This is because it is extremely difficult to learn the kernel in the convolutional layers with a random PMU sequence.

\par After the cross-validation is completed, we train the neural networks with the full training dataset and evaluate their performance on the testing dataset. Specifically, we repeat training and testing for 10  times  with  different  initial neural network weights. The average testing accuracy with respect to training epoch for each method is shown in Fig. \ref{accTest}. The average testing F1 scores for each power system event class is reported in Table \ref{f1score}. The testing results show that \textit{baseline+GSP+info} performs the best in terms of both classification accuracy and F1 scores.
Compared to the baseline model, the combination of GSP based PMU sorting and information loading based regularization work synergistically to boost F1 scores for non-event, line-event, generator event, and oscillation event by 7.3\%, 0.4\%, 4.0\%, and 6.7\% respectively. 
The most dramatic performance improvement can be observed for non-events and oscillation events when GSP based PMU sorting method is applied. In addition, the information loading based regularization is more effective when GSP based PMU sorting is adopted.
\renewcommand{\arraystretch}{1}
\begin{table}[!t]
	\setlength{\tabcolsep}{4.5pt}	
	\caption{F1 Scores for Different Event Classes} 
	    \centering
		\begin{tabular}{ccccc}
			\toprule
			    & \multirow{2}{*}{\textit{Non-event}}  & \multirow{2}{*}{\textit{Line-event}} & \textit{Generator} & \textit{Oscillation} \\ 
			    & & &\textit{event} & \textit{event}\\
			\midrule		
			\textit{Baseline} &  0.885 & 0.966  & 0.895   & 0.911 \\
			\midrule	
			\textit{Baseline+info} &  0.884 & 0.965  & 0.908   & 0.894 \\
			\midrule
			\textit{Baseline+GSP} & 0.928   & 0.976  & 0.904  & 0.944 \\
			\midrule		
			\textit{Baseline+GSP+info} & 0.950   & 0.970  & 0.931  & 0.972 \\
			\bottomrule
		\end{tabular}%
	\label{f1score}%
\end{table} 
\par We also carry out a performance comparison between the proposed approach and two benchmark algorithms that are based on principal component analysis (PCA) and two widely used classifiers: k-nearest neighbors (KNN) and the support vector machine (SVM) \cite{brahma2016real,li2019hybrid}. We denote these two benchmark algorithms as PCA+KNN and PCA+SVM. The PCA is employed to reduce the dimensionality of the $PQ|V|f$ snapshots. We use radial basis function (RBF) as the kernel of SVM. Note that two hyperparameters, \emph{i.e.,} the number of principal components, $N_{PC}$, and the number of nearest neighbors, $K$, need to be determined. We perform the same cross-validation procedures to evaluate the performance of benchmark algorithms under different hyperparameter values. Results show that $N_{PC}=500$ and $K=1$ perform the best for PCA+KNN (Table \ref{f1scoreCrossValKNN}) while $N_{PC}=50$ works the best for PCA+SVM (Table \ref{f1scoreCrossValSVM}). Similarly, we train these two benchmark algorithms with full training data and measure their performance on the testing dataset. The average testing F1 scores of benchmark algorithms for each event class are compared with the proposed approach in Table \ref{f1scoreComparison}. Our \textit{baseline+GSP+info} algorithm clearly outperforms the benchmark algorithms in all different power system event types. It is interesting to notice that both benchmark algorithms achieve significantly higher F1 scores in \textit{Generator-events} than the other types of events.

We also record the training and testing time of different algorithms in this case study (Table \ref{runningTime}). Not surprisingly, the proposed approach needs the most time for training due to its large amount of parameters. Note that the training process is conducted off-line, which does not hinder the proposed approach from online usage. In fact our method can process a new arrival of $PQ|V|f$ snapshot in less than 0.1 seconds.
\renewcommand{\arraystretch}{1}
\begin{table}[!t]
	\setlength{\tabcolsep}{11pt}	
	\caption{Average F1 Scores of PCA+KNN with various $N_{PC}$ and $K$} 
	    \centering
		\begin{tabular}{ccccccc}
			\toprule
			  $N_{PC}$ & 1000 & 800 & 500 & 300 & 200 \\
			\midrule		
			K=1  & 0.513 & 0.547  & 0.594  & 0.584   & 0.536 \\
			\midrule
			K=5  &  0.463 & 0.506  & 0.588  &  0.575  & 0.464 \\
			\midrule
			K=10  & 0.435 & 0.467  & 0.540  & 0.525   & 0.446 \\
			\midrule
			K=50  & 0.372 & 0.382  & 0.398  & 0.433   & 0.369 \\
			\bottomrule
		\end{tabular}%
	\label{f1scoreCrossValKNN}%
\end{table} 
\renewcommand{\arraystretch}{1}
\begin{table}[!t]
	\setlength{\tabcolsep}{10.5pt}	
	\caption{Average F1 Scores of PCA+SVM with various $N_{PC}$} 
	    \centering
		\begin{tabular}{ccccccc}
			\toprule
			  $N_{PC}$ & 200 & 100 & 50 & 20 & 10 \\
			\midrule		
			F1 score  & 0.475 & 0.579  & 0.638  & 0.633   & 0.608 \\
			\bottomrule
		\end{tabular}%
	\label{f1scoreCrossValSVM}%
\end{table} 
\renewcommand{\arraystretch}{1}
\begin{table}[!t]
	\setlength{\tabcolsep}{4pt}	
	\caption{F1 Score Comparison between the Proposed Approach and the Benchmark algorithms} 
	    \centering
		\begin{tabular}{ccccc}
			\toprule
			    & \multirow{2}{*}{\textit{Non-event}}  & \multirow{2}{*}{\textit{Line-event}} & \textit{Generator} & \textit{Oscillation} \\ 
			    & & &\textit{event} & \textit{event}\\
			\midrule		
			\textit{PCA+KNN} &  0.510 & 0.462  & 0.892   & 0.581 \\
			\midrule	
			\textit{PCA+SVM} &  0.383 & 0.602  & 0.884   & 0.643 \\
			\midrule		
			\textit{Baseline+GSP+info} & 0.950   & 0.970  & 0.931  & 0.972 \\
			\bottomrule
		\end{tabular}%
	\label{f1scoreComparison}%
\end{table}
\renewcommand{\arraystretch}{1}
\begin{table}[!t]
	\setlength{\tabcolsep}{7.5pt}	
	\caption{Training and Testing Time Comparison} 
	    \centering
		\begin{tabular}{ccc}
			\toprule
			  & Training time & Testing time (per sample)  \\
			\midrule		
			PCA+KNN  & 36.41s & 0.267s  \\
			\midrule		
			PCA+SVM  & 12.55s & 0.038s  \\
			\midrule		
			\textit{Baseline+GSP+info}  & 3.283h &  0.084s \\
			\bottomrule
		\end{tabular}%
	\label{runningTime}%
\end{table}

\par
\begin{figure}[!t]
	\centering
	\includegraphics[width=0.45\textwidth]{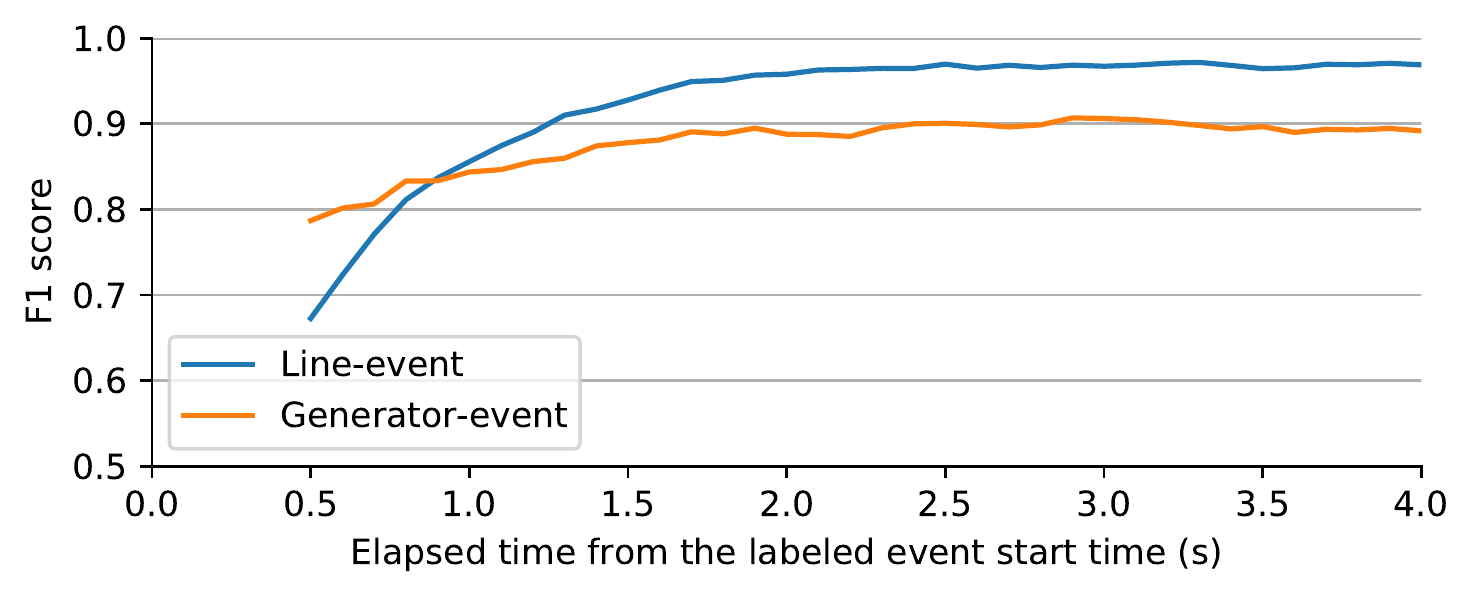}
	\DeclareGraphicsExtensions.
    \caption{Average F1 scores on the \textit{Line-event} and \textit{Generator-event} as the sliding window passes the event start time.}
	\label{accVStimeF1score}
\end{figure}

 We also evaluate how the event classification accuracy changes as the elapsed time from the labeled event start time increases. Our proposed algorithm is expected to achieve higher accuracy after a larger component of the event signature is revealed. In this test, a 12-second sliding window is introduced to simulate the streaming data environment. Initially, the right end of the sliding window is placed at 0.5 seconds passing the labeled event start time. 
It then moves rightward on the $PQ|V|f$ tensor with a step size of 0.1 seconds. We did not start from the exact labeled event start time due to the small differences between the actual event start time stamps and the labeled ones. As shown in Fig. \ref{accVStimeF1score}, the average F1 scores on the \textit{Line-event} and \textit{Generator-event} reach 0.87 and 0.85 in just one second and are saturated two seconds after the event started. Note that oscillation events are excluded because oscillations span the whole time horizon for all available samples.

\subsection{Representation Learning Results}
\begin{figure}[!t]
    \centering
    \subfloat[\textit{Baseline}.]{\includegraphics[width=0.241\textwidth]{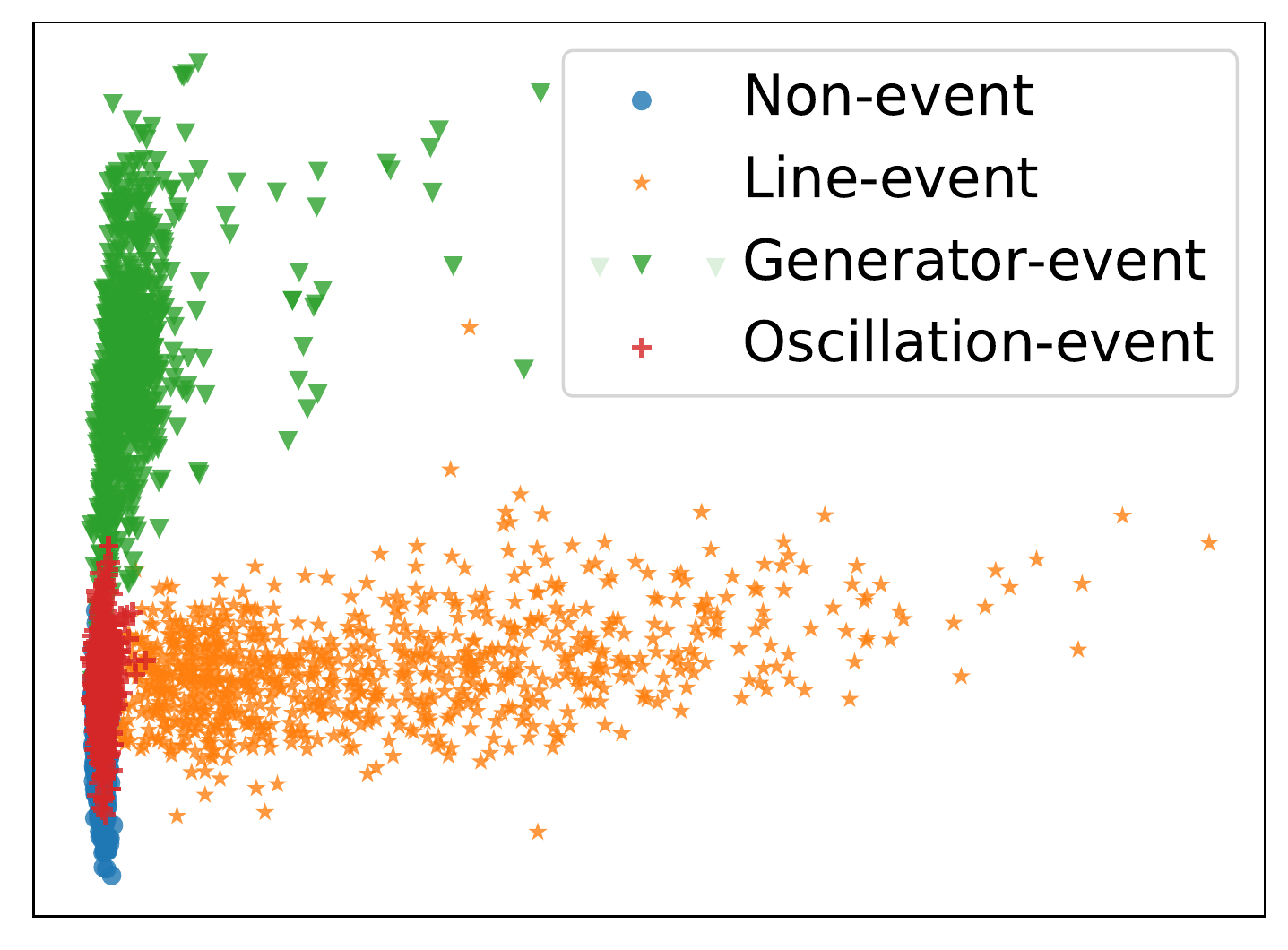}%
    \label{repPlotResNoSort}}
    \hfil
    \centering
    \subfloat[\textit{Baseline+info}.]{\includegraphics[width=0.233\textwidth]{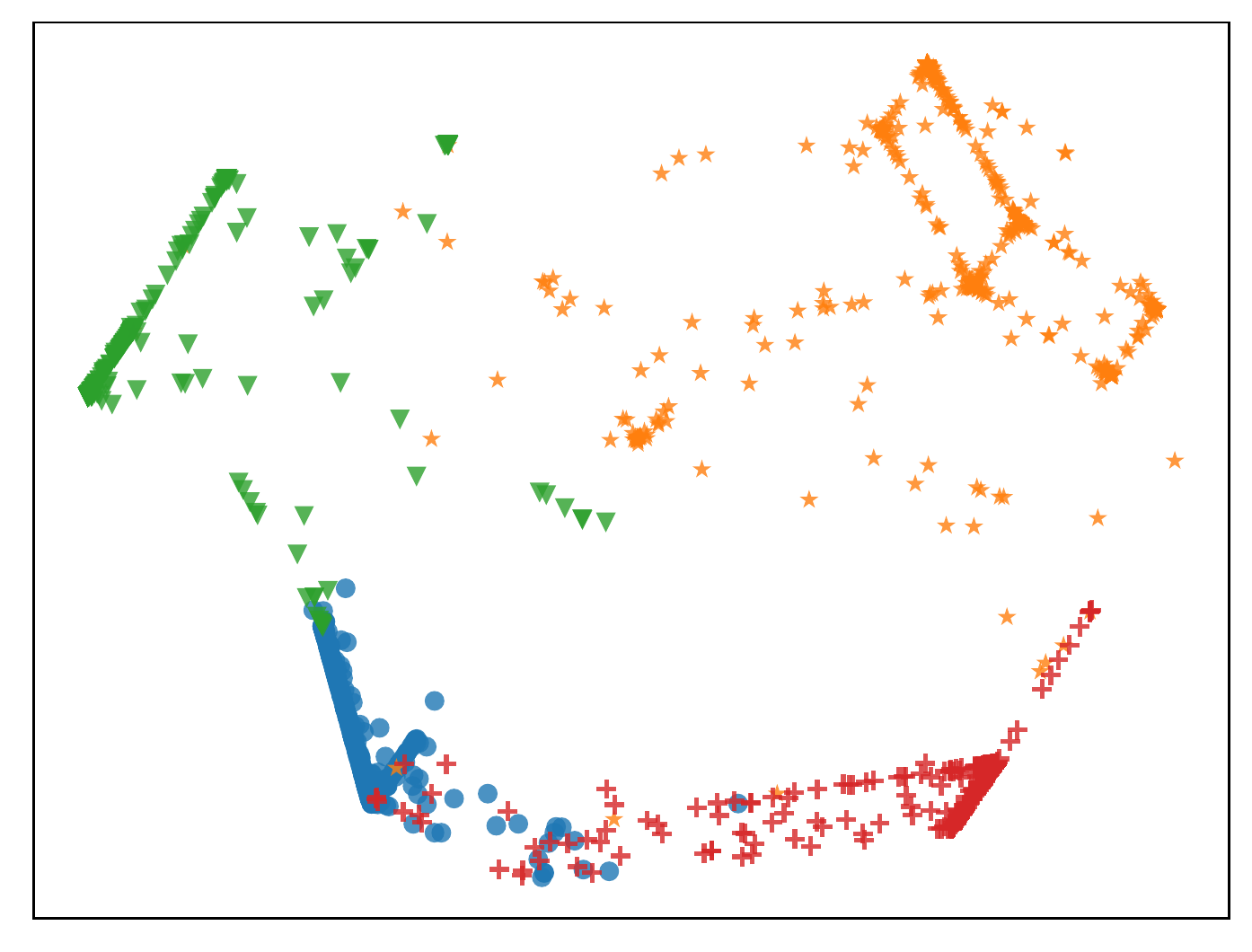}%
    \label{repPlotInfoNoSort}}
    \hfil
    \centering
    \subfloat[\textit{Baseline+GSP}.]{\includegraphics[width=0.241\textwidth]{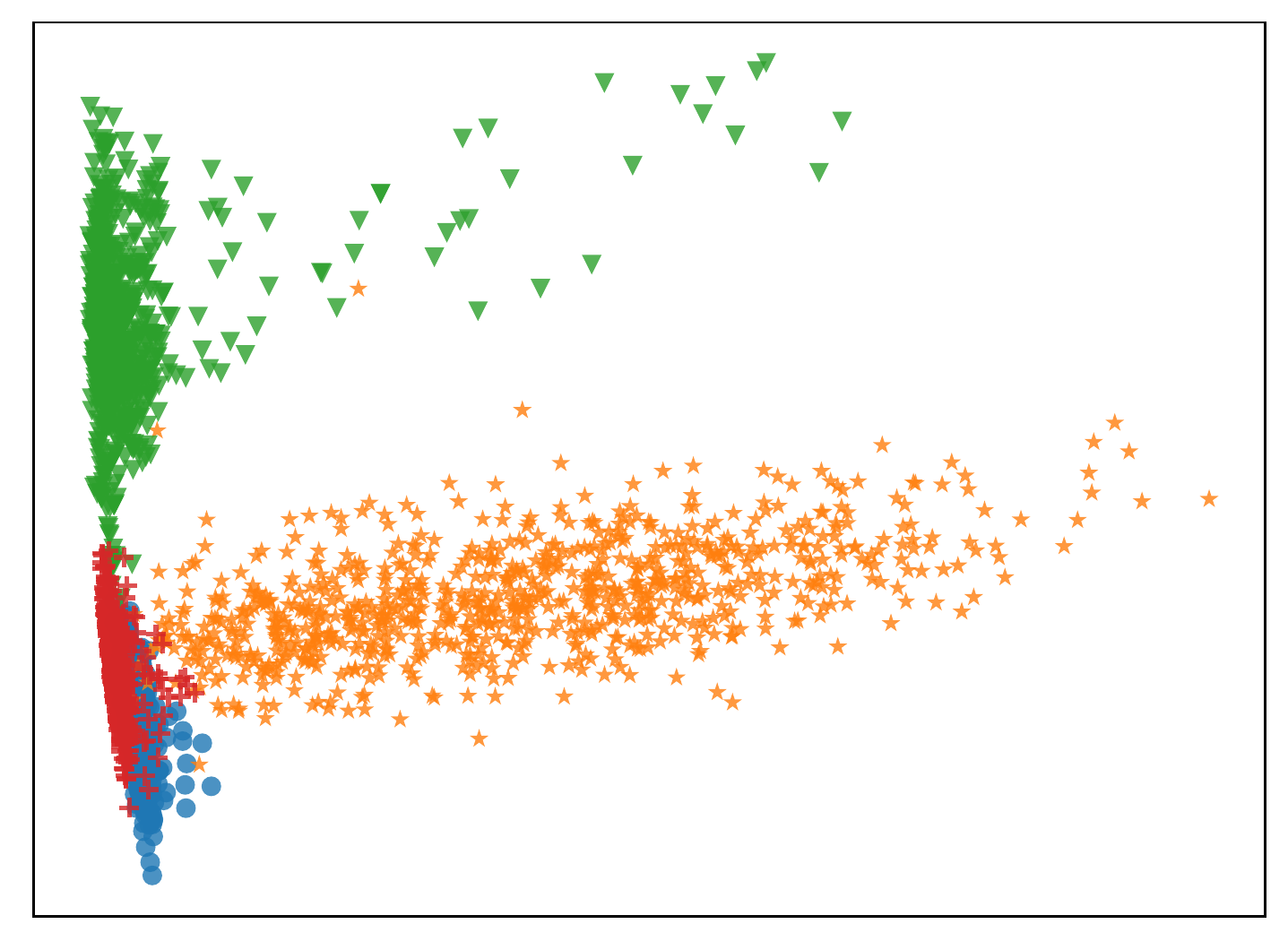}%
    \label{repPlotResSort}}
    \hfil
    \centering
    \subfloat[\textit{Baseline+GSP+info}.]{\includegraphics[width=0.233\textwidth]{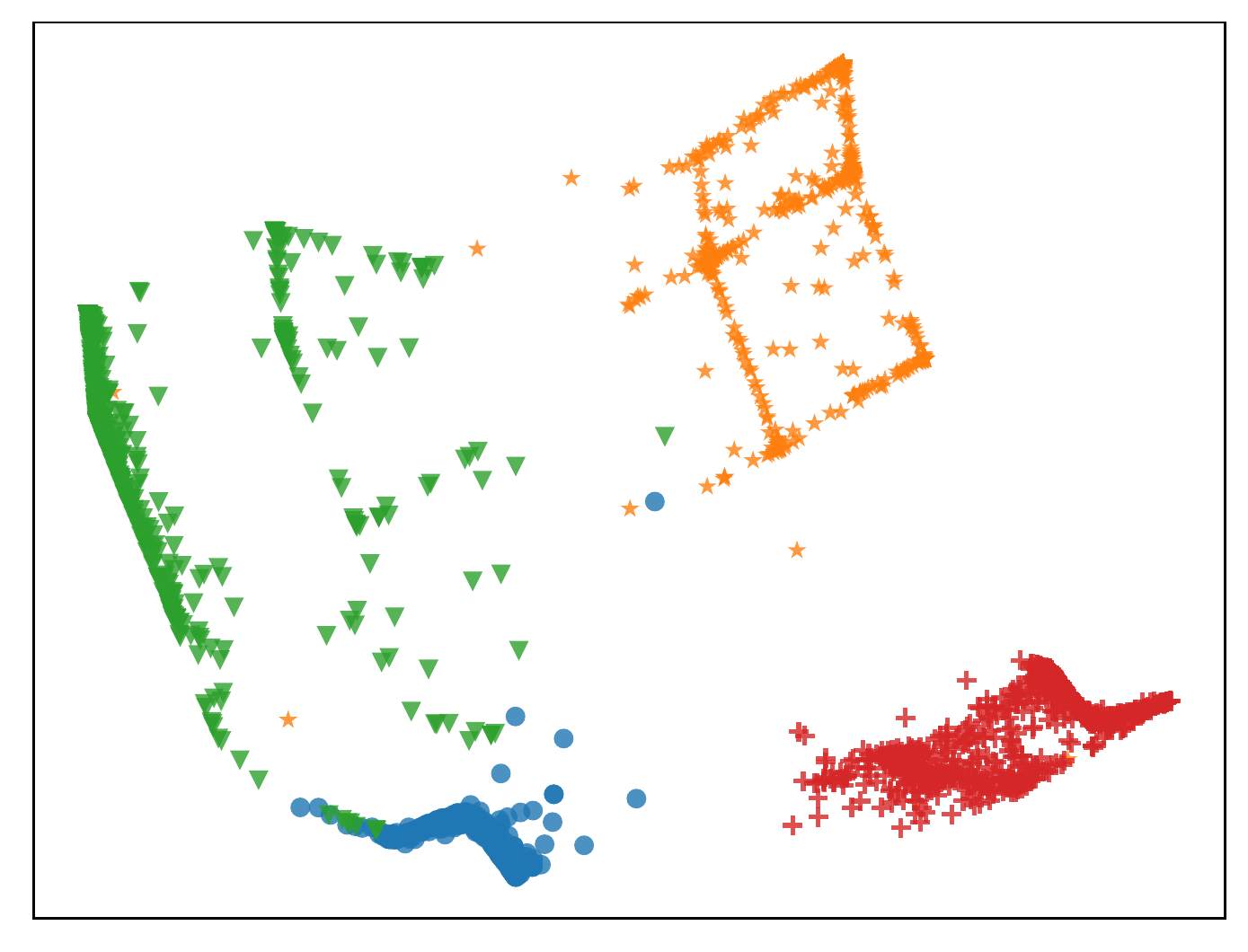}%
    \label{repPlotInfoSort}}
    \caption{Comparison of representations produced by different methods after PCA based dimension reduction.}    
    \label{representationResults}
\end{figure}

The performance of a classifier is primarily dependent on the quality of representations produced by its encoder. The intrinsic goal of GSP based PMU sorting and information loading based regularization is to help encoders learn better representations. Furthermore, representations with higher quality provide better interpretability for the corresponding deep neural network. In this subsection, we visualize and compare the representations learned by different methods.
\par Direct visualization of representations is difficult due to the high dimensionality of the encoders' outputs. To address this issue, researchers typically adopt linear dimensionality reduction methods such as PCA to reduce the dimensionality of the representations to 2 dimensions. Specifically, the principal components are determined through eigendecomposition of sample covariance matrix derived from the representation samples. Then, the original representation data points are projected onto the first two principal components, creating a 2D data array.
\par It is worth noting that linear dimension reduction techniques are preferred over 
nonlinear ones to visualize hidden representations. This is because the hidden representation layer is often followed immediately by a fully connected layer activated by the softmax function. The softmax regression, which serves as the estimator, is essentially a generalized linear separation model. In order for the estimator to achieve great classification performance, the representations should be almost linear separable. Hence, a linear projection such as PCA is usually introduced to visualize the learned representations of the deep neural networks.

\par The dimension reduced hidden representations produced by different methods are shown in Fig. \ref{representationResults}. By comparing Fig. \ref{repPlotResNoSort} and \ref{repPlotInfoNoSort} with Fig. \ref{repPlotResSort} and \ref{repPlotInfoSort}, we observe that the GSP based PMU sorting significantly improves the encoder's ability in separating oscillation events from non-events. Meanwhile, by comparing the sub-figures on the left hand side with that on the right hand side,  we see a higher separation level is gained from the information loading based regularization. Representations with this abundant separation between classes greatly simplifies the corresponding estimator's classification task, which explains the excellent performance of \textit{Baseline+GSP+info}. 

\begin{figure}[!t]
	\centering
	\includegraphics[width=0.35\textwidth]{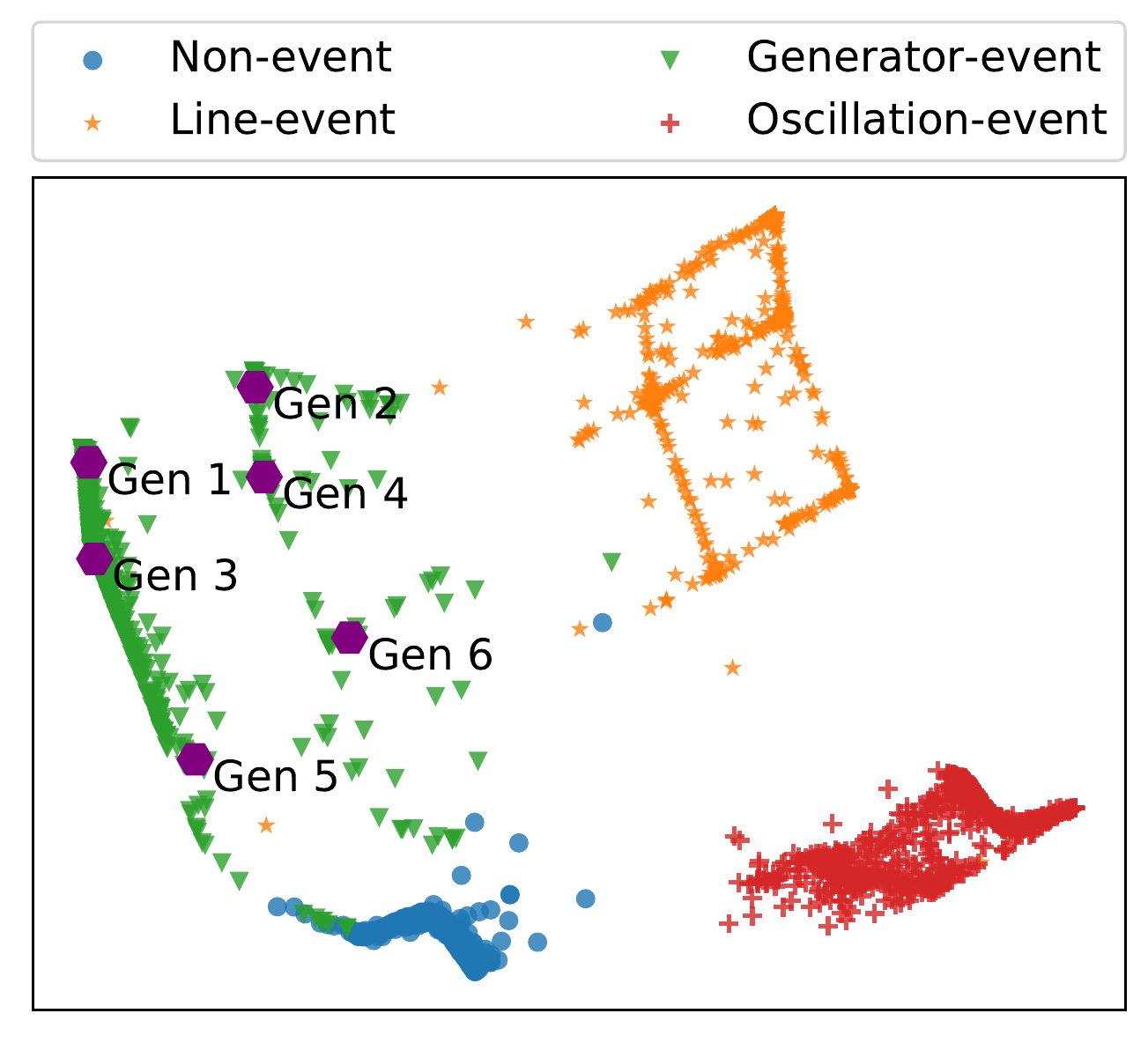}
	\DeclareGraphicsExtensions.
    \caption{Representation locations of six sample generator events.}
	\label{selectedPntsNum}
\end{figure}
\par
\begin{figure}[!t]
	\centering
	\includegraphics[width=0.48\textwidth]{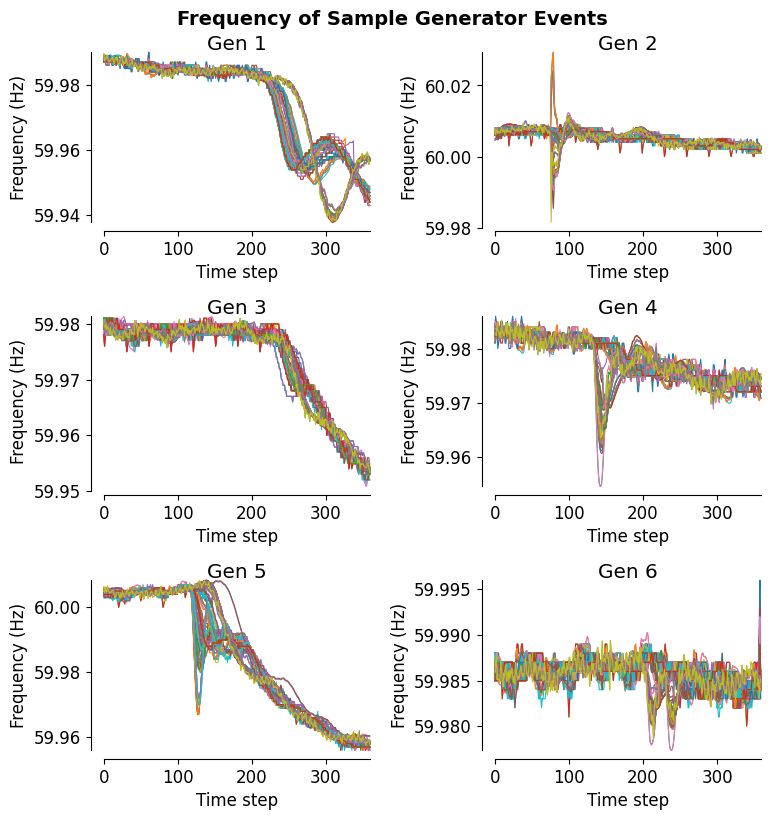}
	\DeclareGraphicsExtensions.
    \caption{Frequency behaviors of six sample generator events.}
	\label{FplotGen}
\end{figure}

 More interestingly, the representations learned by the proposed approach can distinguish power system events with different characteristics. To illustrate this point, we select six generator events with their representations located at different spots (Fig. \ref{selectedPntsNum}). We plot the corresponding frequency ($F$) time series of each sample event in Fig. \ref{FplotGen}. The two green ``lines'' in the representation space represent two clusters of generator events. Generator events 1, 3, and 5 belong to the left hand side (LHS) cluster while generator events 2, 4, and 6 belong to the right hand side (RHS) cluster. As shown in Fig. \ref{FplotGen}, events from the LHS cluster exhibit significantly different frequency behaviors from that of RHS cluster. The decline in frequency of LHS events spans a relatively long time while the frequency drops of RHS events tend to be sharper and shorter. Moreover, the frequency of the power system quickly bounces back to the pre-event level for events from the RHS cluster.

\section{Conclusion} \label{conclusion}
This paper proposes to identify and classify power system events with a deep neural network using streaming PMU dataset. The proposed framework includes three key components: the neural classifier, the GSP based PMU sorting method, and the information loading based regularization. The neural classifier consists of a CNN based encoder and an estimator represented by a dense layer of neurons. The GSP based PMU sorting method places highly correlated PMUs closer to each other, which makes parameter sharing more effective in the CNN based encoder. The information loading based regularization further improves the generalization of the classifier by tuning the mutual information between the input features and the representation. Testing results on large-scale PMU dataset from the Eastern Interconnection of the U.S. transmission grid demonstrate that the proposed approach achieves high accuracy in identifying power system events.

\section*{Acknowledgment}
This material is based upon work supported by the Department of Energy under Award Number DE-OE0000916.



%







\bibliographystyle{IEEEtran}
\bibliography{refs}

\end{document}